\documentclass{article}

\usepackage{arxiv}
\usepackage[utf8]{inputenc} 
\usepackage[T1]{fontenc}    
\usepackage{url}            
\usepackage{booktabs}       
\usepackage{amsfonts}       
\usepackage{nicefrac}       
\usepackage{microtype}      
\usepackage{lipsum}		
\usepackage{graphicx}
\usepackage{adjustbox}
\usepackage{siunitx}
\usepackage{doi}
\usepackage{amsmath,bm,times} 
\usepackage{tikz} 

\usetikzlibrary{matrix}
\usetikzlibrary{shapes, arrows}

\usepackage[numbers]{natbib}

\usepackage{tabularx}
\usepackage{tabularray}
\usepackage{tablefootnote}
\usepackage{booktabs}
\usepackage{float}

\usepackage{hyperref}
\hypersetup{
    colorlinks=true,
    linkcolor=blue,
    filecolor=magenta,      
    urlcolor=cyan,
    pdftitle={Overleaf Example},
    pdfpagemode=FullScreen,
    citecolor=blue,
    }

\title{Predicting Battery Lifetime Under Varying Usage Conditions from Early Aging Data}
\date{}

\author{Tingkai Li$^{1,*}$, Zihao Zhou$^{2,*}$, Adam Thelen$^{3}$, David Howey$^{2}$, Chao Hu$^{1,\dagger}$ \\
\\
	\normalsize $^{1}$Department of Mechanical Engineering, University of Connecticut, Storrs, CT, 06269, USA \\
	\normalsize $^{2}$Department of Engineering Science, University of Oxford, Oxford OX1 3PJ, UK \\
	\normalsize $^{3}$Department of Mechanical Engineering, Iowa State University, Ames, IA 50011, USA \\ 
 \\
	\normalsize $^{*}$Indicates authors contributed equally. \\
	\normalsize $^{\dagger}$Indicates corresponding author. Email: \href{mailto:chao.hu@uconn.edu}{chao.hu@uconn.edu}.
}

\renewcommand{\undertitle}{}

\hypersetup{
pdftitle={Predicting Battery Lifetime Under Varying Usage Conditions from Early Aging Data},
pdfsubject={Early Prediction of Battery Lifetime},
pdfauthor={Tingkai~Li, Zihao~Zhou, Adam~Thelen, David~Howey, Chao~Hu},
pdfkeywords={lithium-ion, battery, lifetime, hierarchical, machine learning, prediction, open data}
}

\begin{document}

\maketitle

\begin{abstract}
Accurate battery lifetime prediction is important for preventative maintenance, warranties, and improved cell design and manufacturing. 
However, manufacturing variability and usage-dependent degradation make life prediction challenging. Here, we investigate new features derived from capacity-voltage data in early life to predict the lifetime of cells cycled under widely varying charge rates, discharge rates, and depths of discharge. Features were extracted from regularly scheduled reference performance tests (i.e., low-rate full cycles) during cycling. The early-life features capture a cell's state of health and the rate of change of component-level degradation modes, some of which correlate strongly with cell lifetime. Using a newly generated dataset from 225 nickel-manganese-cobalt/graphite Li-ion cells aged under a wide range of conditions, we demonstrate a lifetime prediction of in-distribution cells with 15.1\% mean absolute percentage error using no more than the first 15\% of data, for most cells. Further testing using a hierarchical Bayesian regression model shows improved performance on extrapolation, achieving 21.8\% mean absolute percentage error for out-of-distribution cells. Our approach highlights the importance of using domain knowledge of lithium-ion battery degradation modes to inform feature engineering. Further, we provide the community with a new publicly available battery aging dataset with cells cycled beyond 80\% of their rated capacity.

\end{abstract}

\let\thefootnote\relax\footnotetext{\hspace*{-5mm}The battery aging dataset generated for this work will be available for download here: \href{https://doi.org/10.25380/iastate.22582234}{https://doi.org/10.25380/iastate.22582234}.}

\section*{CONTEXT AND SCALE}

Extending the lifetime of lithium-ion batteries is essential for improving their economic and environmental impact. However, measuring battery lifetime can greatly delay product design because cells can sometimes take years to reach their end of life in accelerated laboratory aging tests. Researchers and engineers need quick and easily obtainable cell lifetime diagnostic signals to rapidly validate products and cell designs. Here, we demonstrate a new method for predicting the lifetime of cells operating under widely varying conditions using measurements from early life. These measurements, taken during the first three weeks of testing, quantify a cell's rate of degradation and correlate strongly with lifetime. Our method can be used to predict the lifetime of batteries under a wide range of operating conditions, and could potentially be extended to different chemistries. Although the method requires full-life training data, there are many possible applications for the trained model, such as screening of new cells, or estimates of relative performance between different cell types.

\section{INTRODUCTION}

Understanding the long-term degradation of lithium-ion batteries is crucial for their optimal manufacturing, design, and control \citep{birkl2017degradation, sulzer2021challenge}. However, repeatedly assessing cell performance via aging experiments is a time- and cost-intensive task \citep{thelen2022integrating}. Manufacturers and researchers need quick and accurate methods to screen long-term performance and quantify the impact of new designs and control changes without having to cycle cells to the end of life (EOL) each time a new question arises. Models using data from early life could significantly shorten the time needed to make accurate predictions of long-term degradation \citep{kunz2021early}, and this could lead to rapid screening of new battery performance and optimization of charging protocols \citep{severson2019data, attia2020closed, dave2020autonomous}.

The idea that lifetime can be predicted using measurements from the early stages of battery aging experiments has its roots in research from over a decade ago by J.\ Dahn and researchers at Dalhousie University, who were investigating the impact of new electrolyte additives and electrode designs on battery performance. In late 2009, they published a paper describing how high precision measurements of coulombic efficiency during the first few cycles could be used to predict cell lifetime and rank it qualitatively against other cells \citep{smith2009precision}. Coulombic efficiency is an important performance metric, and it is calculated as the discharge-to-charge capacity ratio, where an ideal value of unity indicates perfect cyclic efficiency. Measuring cell coulombic efficiency with an error of $<$ 0.01\% can indicate cell-to-cell differences caused by different rates of undesirable side reactions that lead to capacity fade. Using purpose-built high precision equipment, the Dalhousie team published a paper in 2011 that compared long-term cycling data ($> 750$ cycles) with predicted lifetimes extrapolated from short-term ($< 500$ hours) high-precision coulombic efficiency measurements \citep{burns2011evaluation}. 

Since this work, many new studies have been published on `early life prediction'. In 2013, the Dalhousie University group published another paper demonstrating the lifetime ranking of 160 Li-ion cells with various electrolyte additives, using high precision coulombic efficiency measurements from the first 50 cycles of data  \citep{burns2013predicting}. The coulombic efficiency measurements strongly correlated with the cells’ lifetimes. However, many researchers and industry professionals do not have access to high precision machines for testing. Furthermore, it would be even more useful to predict lifetimes using early-life measurements made during faster cycling experiments and under a broader range of operating conditions, enabling the technology to be deployed in more research areas and even for cells operating in the field. 

Research by Baumh\"ofer et al.\ \citep{baumhofer2014production} and Harris et al.\ \citep{harris2017failure} investigated alternative approaches not requiring the use of a high precision cycler. Baumh\"ofer et al.\ developed a lifetime prediction model on 48 cells cycled under identical conditions \citep{baumhofer2014production}. Hundreds of early-life features extracted from impedance spectra, pulse characterization tests at different states of charge, and standard capacity tests were reduced to a set of 24 features and used for prediction. The model using 24 features was accurate within 16 cycles, however, further analysis showed that model accuracy was highly dependent on the number of features used, with more features generally being better, suggesting the model may possibly be overfitting the small dataset ($N = 48$). Harris et al.\ examined the failure statistics of 24 cells cycled under identical conditions and established a weak correlation between the cells’ capacity at cycle 80 and the capacity at cycle 500 \citep{harris2017failure}. These works suggest simpler and more easily obtainable early-life features might be found to correlate with eventual lifetime.

Severson et al.\ \citep{severson2019data} in 2019 demonstrated an early life prediction model using features extracted from the discharge capacity vs.\ voltage $(Q(V))$ curves during regular cycling. The feature extraction method was unique, quantifying the cells’ degradation rates by tracking the early-life variation of their $Q(V)$ curves between cycles 10 and 100, referred to as $\Delta Q_{100-10}(V)$. The approach was also used in follow-up work by Attia et al.\ \citep{attia2020closed} to accelerate an experimental campaign to optimize the constant current portion of a fast charging protocol. The researchers in these papers generated a large battery aging dataset from 169 lithium-iron-phosphate/graphite (LFP) cells cycled under various fast charging protocols. This was made publicly available, and many other researchers have investigated methods of further improving predictive performance and feature extraction techniques using this data \citep{greenbank2021automated,zhang2021prognostics, yang2020lifespan, weng2021predicting, saxena2022convolutional, herring2020beep, fei2021early, fermin2020identification, li2021one, paulson2022feature}. Notably, Paulson et al.\ \citep{paulson2022feature} demonstrated accurate early life prediction on six different metal oxide cathode chemistries. Fermin-Cueto et al.\ \citep{fermin2020identification} investigated predicting the knee point (when capacity begins to decrease rapidly) in a cell’s capacity degradation curve using early-life features. Similarly, Li et al.\ \citep{li2021one} demonstrated a prediction model capable of projecting the entire capacity degradation trajectory from early-life features. 

Despite this growing body of research, many fundamental questions about battery life modeling remain unanswered. One fundamental issue is that, in order to train machine learning models to predict lifetime from early-life cycles, data from the \emph{entire} lifetime is required. Therefore these approaches are best suited to applications such as screening cells after manufacturing, or relative comparisons, rather than quantitatively absolute predictions. A second issue is a lack of publicly available battery-lifetime data that covers a wide range of conditions. The dataset published in \citep{severson2019data, attia2020closed} was specifically generated to study high-rate fast charging protocols for LFP cells, leaving the discharge rate and depth of discharge fixed. Even though the dataset is relatively large compared to existing publicly available datasets ($N=169$ cells), the limited range of operating conditions, in this case, induced a single dominant degradation mode (loss of active material at the anode or negative electrode, “$\mathrm{LAM}_{\mathrm{NE}}$”), causing all of the capacity degradation trajectories to have very similar shapes, and perhaps making lifetime prediction easier \citep{attia2022knees}. While the relationships between cell operating conditions and the corresponding degradation modes are well understood \citep{birkl2017degradation, thelen2022integrating, waldmann2018li, han2019review}, it remains unclear how the $\Delta Q(V)$ feature transfers to cells of different chemistries and to situations where multiple interacting degradation modes are present. This is especially the case for cells that experience milder degradation resulting in less obvious changes in the $Q(V)$ curve. Furthermore, all cells in the dataset from \citep{severson2019data, attia2020closed} were cycled under a fixed depth of discharge, making it easy to extract features from any cycle along the cell's degradation trajectory. However, in practice, cells are rarely subjected to full depth-of-discharge cycles, so there is a need to explore alternative methods of collecting early-life feature data and validating results using periodic reference performance tests or other means.

In this work, we investigate new early-life features derived from capacity-voltage data that can be used to predict the lifetimes of cells cycled under a wide range of charge rates, discharge rates, and depths of discharge. To study this, we generated a new battery aging dataset from 225 nickel-manganese-cobalt/graphite cells, cycled in groups of four per condition, under a much wider range of operating conditions than existing publicly available datasets \citep{dos2021lithium}. The cells in our dataset exhibit larger variations in their capacity degradation trajectories than previous open datasets, driven by the interactions and accumulations of various component-level degradation mechanisms \citep{birkl2017degradation, attia2022knees}. To predict the lifetimes of cells experiencing different degradation pathways accurately, we introduce new early-life features extracted from the differential voltage ($dV/dQ$ vs.\ $Q$) and incremental capacity ($dQ/dV$ vs.\ $V$) data gathered during regular weekly reference performance tests (RPTs). The RPTs, two complete cycles at full depth of discharge, enable consistent feature extraction and lifetime prediction for cells that normally cycle at fractional depths of discharge, some as low as 4.0\%. Using as little as the first 5\% of the aging data, we achieve a prediction error of 22\% $\mathrm{MAPE}$ on the lifetime. Including up to 15\% of the entire cell lifetime data, we achieve an average prediction error of 2.8 weeks $\mathrm{RMSE}$ and 15.1\% $\mathrm{MAPE}$ on in-distribution test sets when testing the new features in traditional machine learning models built with regularized linear regression. Given that our dataset has a hierarchical structure (i.e., the `group' level and the `cell' level) in nature, we also explore the possibility of applying hierarchical Bayesian linear modeling to predict lifetime, which achieves better extrapolation performance on out-of-distribution samples, viz.\ 7.3 weeks $\mathrm{RMSE}$ and 21.8\% $\mathrm{MAPE}$ lifetime prediction error.

The major contributions of this work are 
\begin{itemize}
 \item the introduction of a set of new early-life features derived from differential voltage and incremental capacity data,
 \item new approaches for tackling the challenge of feature extraction caused by the wide variation of DoDs,
 \item demonstration of the improvement in accuracy possible using hierarchical Bayesian linear models compared to traditional regression models for lifetime prediction when cells are cycled with varying conditions, and
  \item a large and unique battery aging dataset consisting of 225 NMC cells cycled under a wide range of operating conditions, enabling researchers without access to battery testing equipment to study lifetime modeling.

\end{itemize}

\section{RESULTS and DISCUSSION}
\label{sec:results_and_discussion}

\begin{figure}
	\includegraphics[width=1.0\textwidth]{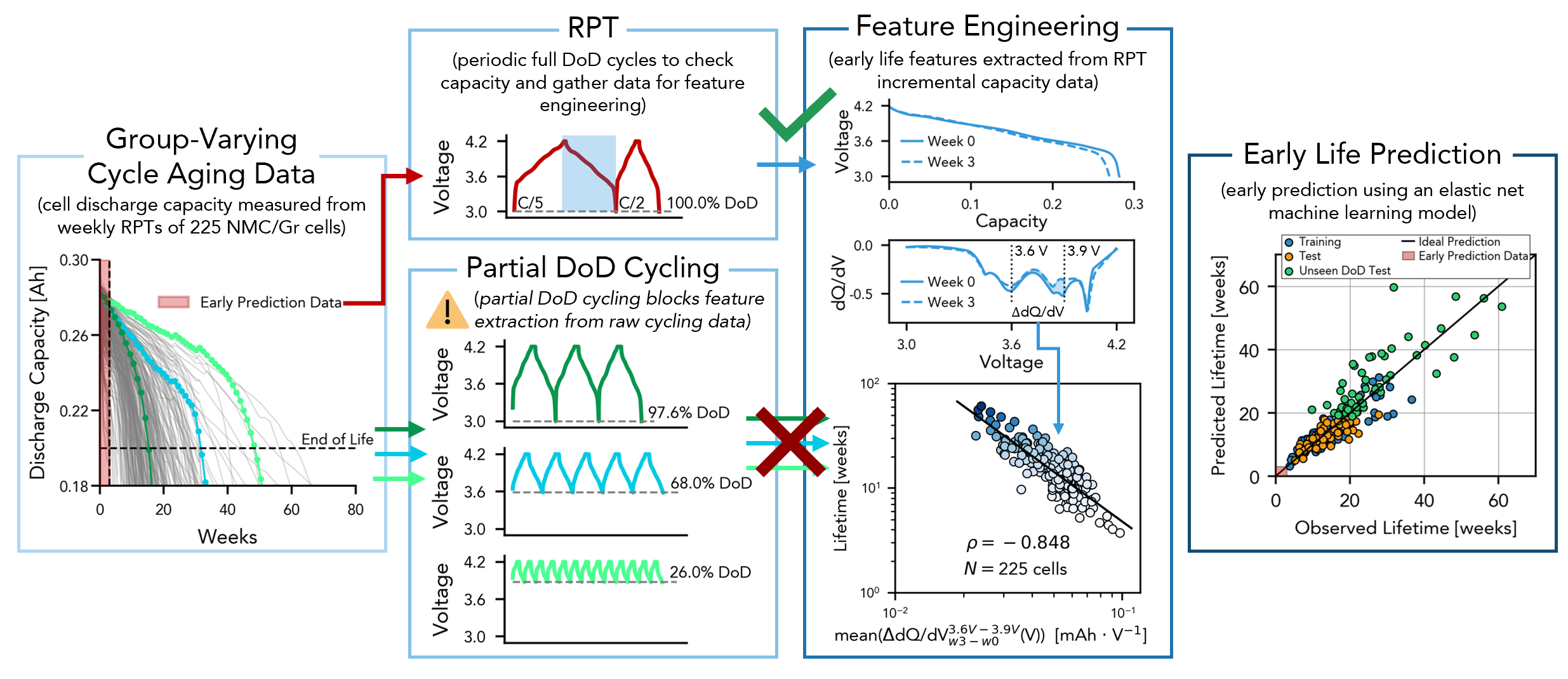}
	\centering
	\caption{High-level overview of our approach. Unlike existing approaches for early prediction, we extract features from periodic reference performance tests instead of regular cycling data. In this example, we extract a feature from a partial voltage window of incremental capacity that is highly correlated with lifetime. From this and other features, we build a machine learning model to predict the lifetimes of new unseen cells.}
	\label{fig:methodology_overview}
	\centering
\end{figure}

Prediction of lifetime from early data is more challenging when there are multiple varying stress factors, because this leads to diverging capacity trajectories. Our approach, outlined in Fig.\ \ref{fig:methodology_overview}, differs from the prior art \citep{burns2011evaluation, burns2013predicting, smith2009precision,severson2019data} in several ways. %
First, to apply early prediction to cells cycled under different depths of discharge, we extract features from periodic RPTs instead of regular cycling data. This means that the discharge voltage curves obtained from periodic RPTs are complete and consistent for every cell, making feature extraction more consistent. Second, we develop new features based on partial voltage windows of $Q(V)$ curves and their derivatives (differential voltage and increment capacity data). Using a new feature extraction method(see details in Sec. \ref{sec:ica_feature_extraction}), we find features that better correlate with cell lifetime for our dataset than existing features reported in the literature \citep{severson2019data, yang2020lifespan, fei2021early}. Additionally, we explore using cycling protocol information ($\mathrm{C}_\mathrm{chg}/\mathrm{C}_\mathrm{dis}/\mathrm{DoD}$) as features to predict lifetime, establishing a link between the two. All extracted features are reduced to a highly predictive subset using a feature selection method (see Sec. \ref{sec:feature_selection}). Then, the selected features are used as input to a machine learning model to predict cell lifetime. In what follows, we outline our approach to feature engineering for early life prediction and discuss the challenges of applying existing feature engineering methodologies proven on LFP/Gr to our NMC/Gr cells that are cycled under a wider range of operating conditions. Last, we introduce hierarchical Bayesian models for early life prediction. %

\subsection{Lithium-ion Battery Dataset Under Varying Usage Conditions}

Publicly available datasets such as those from NASA \citep{bole2014adaptation, saha2008prognostics}, CALCE \citep{he2011prognostics, xing2013ensemble}, and Sandia National Lab \citep{preger2020degradation} contain cells of different chemistries cycled under a range of charge rates, discharge rates, and temperatures. These datasets are frequently used in research studies since they comprehensively report capacity, internal resistance (NASA and CALCE), voltage, current, and temperature. However, the relatively small size of these datasets (roughly 30 cells per group) makes investigating machine learning-based approaches to early life prediction challenging. On the other hand, datasets such as those from the Toyota Research Institute \citep{severson2019data, attia2020closed} and Argonne National Lab \citep{paulson2022feature} contain many more cells (> 150 cells). However, they focus on a limited range of operating conditions---fast charging and symmetric C/2 cycling, respectively---making it difficult to build machine learning models that generalize across cycling conditions.

\begin{figure}
	\includegraphics[width=\textwidth]{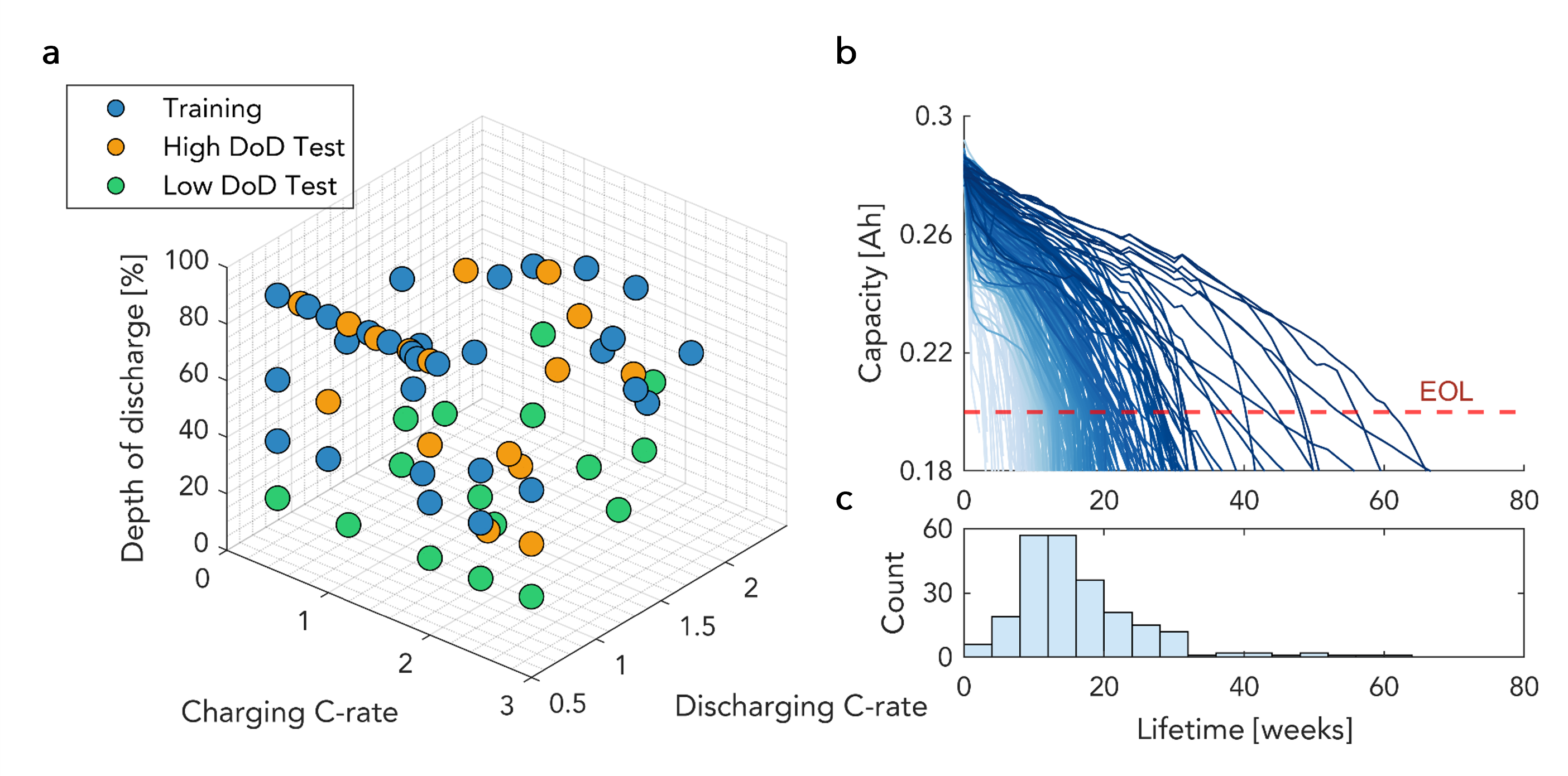}
	\centering
	\caption{Overview of battery aging test conditions and capacity data. \textbf{a}, 3D scatter plot showing train-test split and cycling conditions used -- each point represents conditions for a group of four cells, and marker color indicates a data subset used to generate prediction results in Sec.\ \ref{sec:results_and_discussion}. \textbf{b}, Discharge capacity fade curves for all 225 NMC/graphite cells plotted past 80\% their rated capacity (\SI{250}{mAh}); color of each curve is scaled by cell lifetime. \textbf{c}, Histogram of the cell lifetimes at end-of-life (EOL) using 80\% of rated capacity as threshold.}
	\label{fig:cap_fade}
	\centering
\end{figure}

In light of this, we designed our battery aging dataset to study more cells under a broader range of operating conditions than current publicly available datasets \citep{dos2021lithium}. Our dataset comprises 225 cells cycled in groups of four to capture some of the intrinsic cell-to-cell aging variability \cite{dechent2021estimation}. A unique feature of our dataset is the many capacity degradation trajectories that reflect different accumulated degradation modes induced by the various operating conditions. These trajectories, shown in Fig.\ \ref{fig:cap_fade}, exhibit different one-, two-, and three-stage degradation trends driven by the interaction and accumulation of hidden, threshold, and snowballing degradation modes \citep{attia2022knees}. These varying trends produce cell lifetimes from 1.5 to 60.9 weeks. Experimental details and testing procedures used to generate the dataset can be found in Sec. \ref{sec:test_design} and Supplementary Information. %

\subsection{Extracting Predictive Features from Early Usage Data}
\label{sec:feature_extraction}
\begin{figure}
	\includegraphics[width=.7\textwidth]{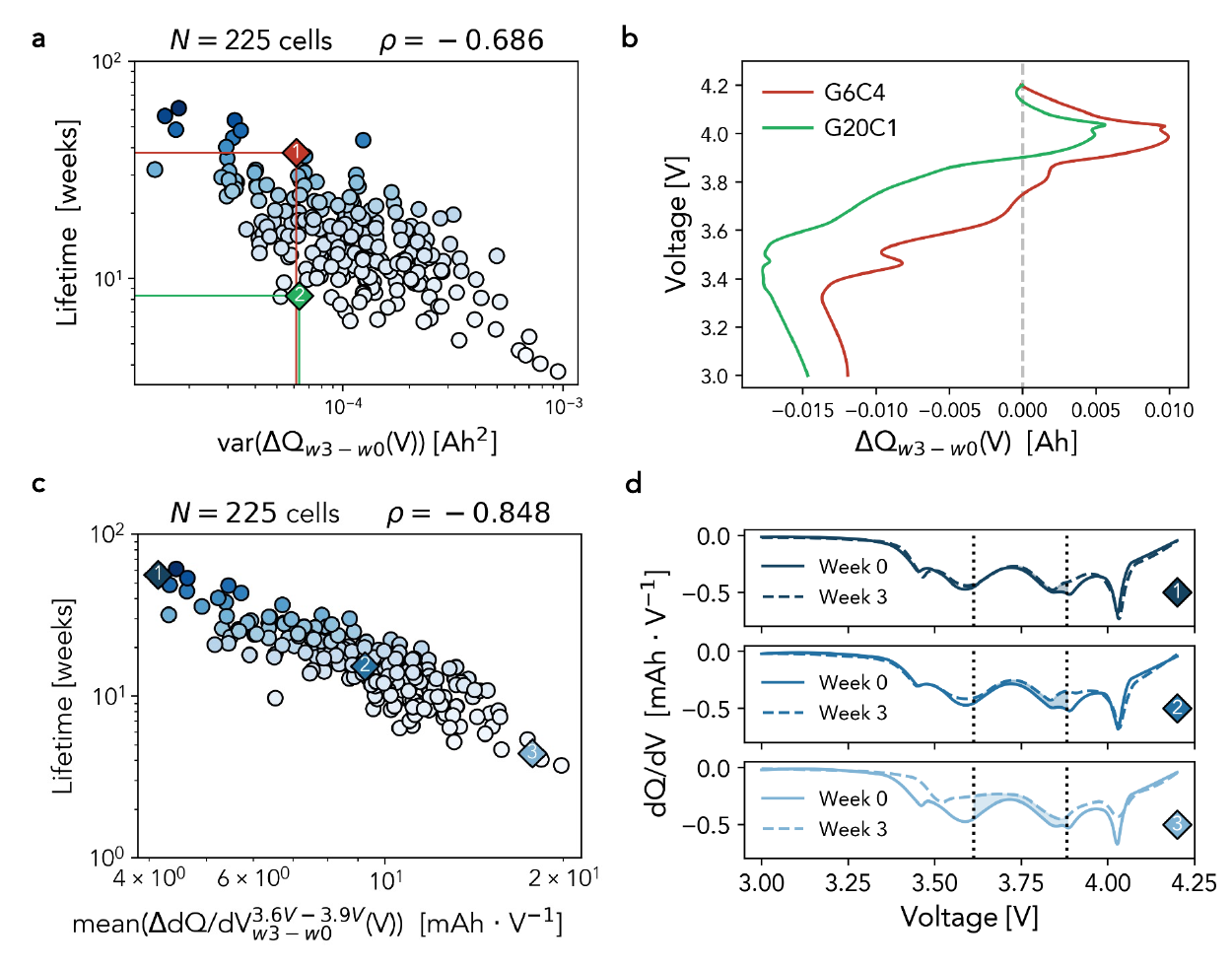}
	\centering
	\caption{Well-known early-life features do not explain the variance in our dataset, and a newly extracted feature from incremental capacity curves correlates better with lifetime. \textbf{a}, Cell lifetime for 225 NMC cells plotted as a function of $\mathrm{var}(\Delta Q_{\mathrm{w3}-\mathrm{w0}}(V))$; Pearson correlation coefficient -0.686. The two cells highlighted have similar values of $\mathrm{var}(\Delta Q_{\mathrm{w3}-\mathrm{w0}}(V))$ but very different lifetimes. \textbf{b}, Difference between discharge capacity curves as a function of voltage between week three and zero for the two cells highlighted in \textbf{a}. \textbf{c}, Cell lifetime plotted as a function of optimized feature $\text{mean}(\Delta dQ/dV_{\mathrm{w3}-\mathrm{w0}}^{3.60\mathrm{V}-3.90\mathrm{V}}(V))$, Pearson correlation coefficient $-0.848$. \textbf{d}, Incremental capacity curves from weeks three and zero for three representative cells; the change in these between the voltage limits over the first three weeks is shaded. }
	\label{fig:f_eng}
	\centering
\end{figure}

Initially, we extracted features previously reported to correlate strongly with cell lifetime \citep{severson2019data, yang2020lifespan, fei2021early}. We adopt the notation $\Delta Q_{\mathrm{w3}-\mathrm{w0}}(V)$ to describe the features, where the subscripts $\mathrm{w3}$ and $\mathrm{w0}$ correspond to data obtained from the RPTs from weeks three and zero, respectively. Preliminary testing of these well-established early-life features reveals that they do not fully explain the variance in our dataset. This is illustrated in Fig.\ \ref{fig:f_eng}a, where we extract the $\mathrm{var}(\Delta Q(V))$ feature reported by Severson et al.\ \citep{severson2019data} using discharge data from RPTs $\mathrm{var}(\Delta Q_{\mathrm{w3}-\mathrm{w0}}(V))$ and plot it against lifetime, revealing a large unexplained variance in the predicted lifetimes. %

To understand why this occurs, consider two cells (G6C4 and G20C1) that have similar feature values but vastly different lifetimes. In this case, even though the $\Delta Q(V)$ curves have the same variance, they do not have the same shape and location (Fig.\ \ref{fig:f_eng}b). 
It can be seen that the group twenty cell (G20C4) experienced more significant capacity loss during this time, evident by the endpoint of $\Delta Q(V)$ at 3.0 \SI{3.0}{V}. 
Other noticeable changes exist in the $dV/dQ(Q)$ curves that differ between the cells (shown in Supplementary Information), indicating additional but more subtle degradation modes are present. However, these differences in the evolution of the $Q(V)$ curve during early life are not captured by the feature $\mathrm{var}(\Delta Q_{\mathrm{w3}-\mathrm{w0}}(V))$, causing the unexplained variance in the dataset.

While we only showed an example in Fig.\ \ref{fig:f_eng} for this particular feature, $\mathrm{var}(\Delta Q_{\mathrm{w3}-\mathrm{w0}}(V))$, the unexplained variance in the data persists using most other early-life features we tested. Typically, it is not a requirement that all model input features exhibit a strong correlation with cell lifetime, but finding a few features that do correlate well is generally advantageous because it can improve model fit and accuracy. In light of this, we explored extracting features from differential voltage and incremental capacity curves using partial voltage ranges in order to capture the diverse degradation trends observed in our dataset more accurately. 

To engineer predictive features from incremental capacity ($dQ/dV (V)$) curves, we employed a grid search method to find a voltage range where extracted features, specifically the statistics of curve differences, correlate highly with lifetime (see Sec. \ref{sec:ica_feature_extraction} for more details). 
We find the voltage range that produces the highest linear correlation with cell lifetime is a mid-range where the upper and lower voltage limits are centered around prominent peaks in the incremental capacity curves at ~\SI{3.60}{V} and ~\SI{3.90}{V}. Fig.\ \ref{fig:f_eng}c shows that the change in incremental capacity in this range is inversely proportional to lifetime. This new feature shows a much stronger correlation with cell lifetime and better explains the variance in our dataset compared with the traditional feature $\mathrm{var}(\Delta Q_{\mathrm{w3}-\mathrm{w0}}(V))$. %

This new feature likely captures the rate of active material loss during early life. This idea is supported by degradation diagnostics literature which shows that changes in the intensity of the incremental capacity (\unit{mAh \per V}) curve at constant voltage correspond to a loss of active material \citep{birkl2017degradation, pastor2017comparison, berecibar2016degradation1, berecibar2016degradation2}. The new feature captures the change in incremental capacity intensity, calculated as the mean change in \unit{mAh \per V} over the middle voltage range, $\mathrm{mean}\left( \Delta dQ/dV_{\mathrm{w3}-\mathrm{w0}}^{3.60\mathrm{V}-3.90\mathrm{V}}(V)\right) = \mathrm{mean}\left(dQ/dV_{\mathrm{w3}}^{3.60\mathrm{V}-3.90\mathrm{V}}(V) - dQ/dV_{\mathrm{w0}}^{3.60\mathrm{V}-3.90\mathrm{V}}(V)\right)$, see Fig.\ \ref{fig:f_eng}d. Additional analysis to understand this feature regarding degradation information has been demonstrated using experimental half-cell data and included in the Supplementary Information.

Lifetime modeling work on NMC/Gr cells by Smith et al.\ \citep{smith2021lithium} showed that the capacity fade rate due to cycling tracked nearly linearly with the square-root-of-cycling throughput, calculated as $({\mathrm{C}_{\mathrm{chg}}}\mathrm{DoD})^{0.5}$, where $\mathrm{C}_{\mathrm{chg}}$ is charging C-rate and DoD is depth of discharge for the experiments. This metric is described as tracking the concentration gradient of lithium ions in the cathode active material and is a proxy for diffusion-induced stress \citep{smith2021lithium, reniers2019review, smith2006solid}. We further investigate this feature as a model input for early-life prediction (Sec. \ref{sec:stress_features}) and as a condition-level grouping variable for our hierarchical Bayesian modeling approach (Sec.\ \ref{sec:hbm_for_early_prediction}).

The remaining features extracted from incremental discharge capacity curves are based on the previously identified voltage range of 3.60 $-$ \SI{3.90}{V}. We use the upper and lower voltage limits imposed during cycling to create two more ranges, 3.00 $-$ \SI{3.60}{V} and 3.60 $-$ \SI{4.20}{V}. We then extract two features from each voltage range using the mean and variance summary statistics. In total, we extracted six features from $\Delta dQ/dV(V)$, two from each of the three voltage ranges using the mean and variance summary statistics.

\subsection{Partitioning Data for Machine Learning and Feature Selection}

\begin{figure}
	\includegraphics[width=1\textwidth]{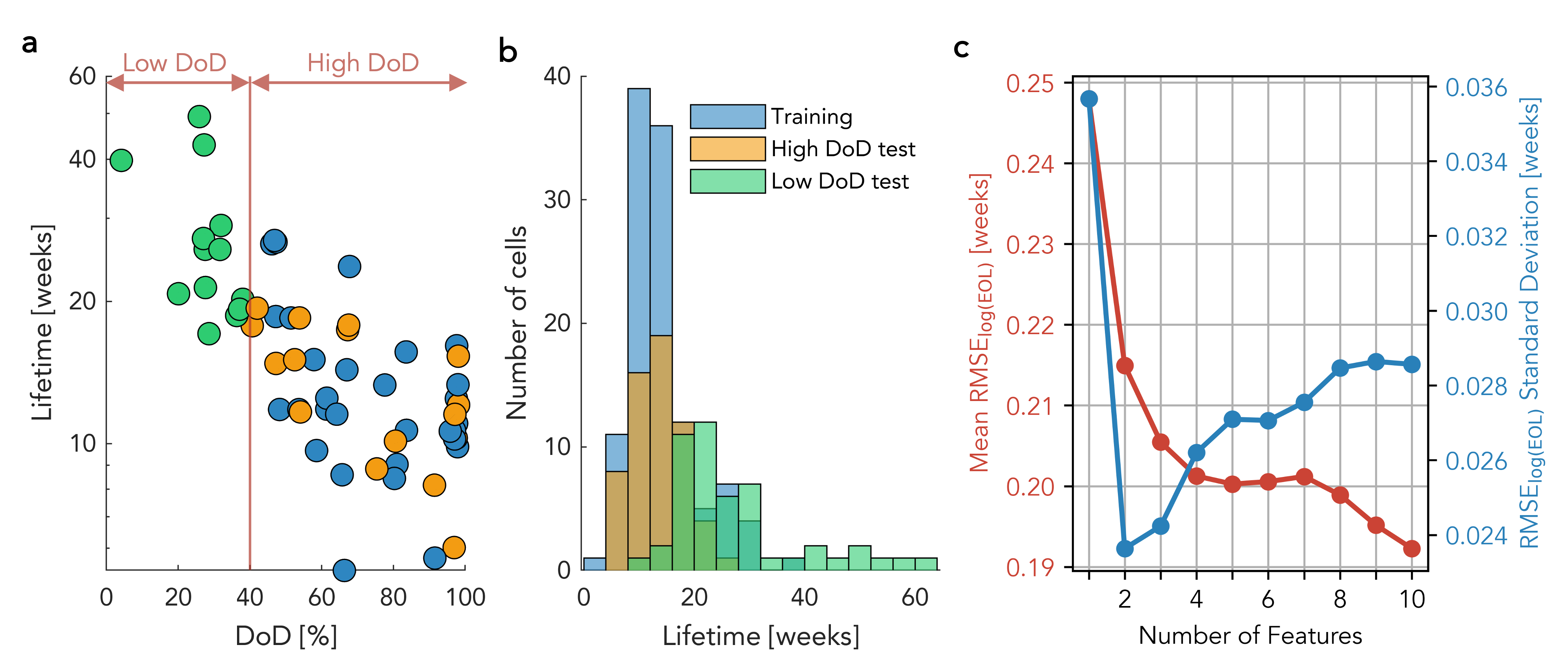}
	\centering
	\caption{\textbf{a.} Scatter plot of mean group lifetime vs.\ DoD; marker color indicates train/test subset. \textbf{b.}  Histogram showing each subset's distribution of cell lifetimes. \textbf{c.} Mean and standard deviation of $\mathrm{RMSE_{\log(EOL)}}$ for five-fold repeated cross-validation on the ten candidate models.}
	\label{fig:partition}
	\centering
\end{figure}

Dataset partitioning was done at the group rather than the cell level, for three reasons. First, practical battery aging tests for product validation typically cycle multiple cells under the same conditions to capture the aging variability due to manufacturing. Second, it is desirable to build an early prediction model to predict the lifetimes of cells cycled under previously untested conditions. Finally, although building an early prediction model with cells tested under rapidly accelerated aging conditions is useful in minimizing the time and costs of collecting aging data, one cannot preemptively know the lifetime (before tests), so grouping must be done using an alternative indicator of cell lifetime. Since the depth of discharge is the dominant cycling stress factor impacting the battery lifetimes in our aging dataset (Fig.\ \ref{fig:partition}a), this was used to determine the dataset partitioning. 

We first separate our dataset into a high-DoD region and a low-DoD region, with a boundary at 40\% depth of discharge (Fig.\ \ref{fig:partition}a). In the high-DoD region, we further divide the data into a training set and an in-distribution high-DoD test set. The high-DoD test set is used to evaluate the model's prediction accuracy for cells with conditions similar to the ones the model was trained on. Last, we assign all data in the low-DoD region ($< 40\%$) to a second test set used to test the model's ability to extrapolate to unseen test conditions. The dataset split is also visualized in Fig.\ \ref{fig:cap_fade}a, where each axis is one of the three cycle aging stress factors ($\mathrm{C}_\mathrm{chg}/\mathrm{C}_\mathrm{dis}/\mathrm{DoD}$), and the marker color indicates the data subset that the group belongs to. The training set contains cells with lifetimes ranging from 3.7 to 36.6 weeks, and the high-DoD test set has cells with lifetimes between 5.2 and 31.6 weeks. On the other hand, the low-DoD test set is more diverse, with lifetimes ranging from 9.7 to 60.9 weeks. Histograms of cell lifetimes for each data subset are visualized in Fig.\ \ref{fig:partition}b.

After extracting the features outlined in Sec.\ref{sec:feature_extraction} and Supplementary Information, we perform feature selection on the training dataset following the method described in Sec.\ \ref{sec:feature_selection}. All extracted features are outlined in the Supplementary Information. To avoid poor performance on the test datasets due to over-fitting, we perform a study of five repeated five-fold cross-validation using up to 10 features.

Repeated cross-validation is intended to minimize the statistical randomness caused by a single five-fold cross-validation partition. The trends of the mean and standard deviation of cross-validation  $\mathrm{RMSE}_{\mathrm{EOL}}$ of this trial are reported in Fig.\ \ref{fig:partition}c, and the selected feature in each step is listed in Table \ref{table:CV_selection_result}. The model with two features, namely $\log \left(\mathrm{mean}(\Delta dQ/dV_{\mathrm{w3}-\mathrm{w0}}^{3.6V-3.9V}(V)\right)$ and $\log \left(\left|\Delta \mathrm{CV\; Time}_{\mathrm{w3}-\mathrm{w0}})\right|\right)$, has the lowest run-to-run variance and relatively low mean error $\mathrm{RMSE_{EOL}}$. Adding a third feature to the set, $\mathrm{DoD}$, produces a model with lower mean $\mathrm{RMSE_{EOL}}$ but increases the run-to-run variation. For a more comprehensive evaluation, we compare the results of models trained using both two and three features.

\renewcommand{\arraystretch}{1.5}
\begin{table}
\label{table:CV_selection_result}
\centering
\caption{Step-wise Forward Search Results}
\begin{adjustbox}{max width=\textwidth}

\begin{tabular}{llll} 
\toprule
\textbf{Step Number} & \textbf{Selected Feature} & \textbf{Description}\\ 
\hline
1 & $\log (\mathrm{mean}(\Delta dQ/dV_{\mathrm{w3}-\mathrm{w0}}^{3.6V-3.9V}(V))$ & Best incremental capacity feature from Sec.\ \ref{sec:ica_feature_extraction} Fig.\ \ref{fig:f_eng}c \\
2 & $\log (\left|
\Delta \mathrm{CV\; Time}_{\mathrm{w3}-\mathrm{w0}}\right|)$ & Change in CV hold time (see Supplementary Information)\\
3 & $\mathrm{DoD}$ & Depth of discharge\\
4 & $\Delta Q_{\mathrm{w3}-\mathrm{w0}}^{1}$ & Change in DVA-based capacity $Q^{\mathrm{DVA,1}}$ (see Supplementary Information)\\
5 & ${\mathrm{C}_{\mathrm{chg}}}^{0.5}\mathrm{DoD}^{0.5}$ & Charge-induced stress (see Sec. \ref{sec:stress_features})\\
6 & $\mathrm{C}_{\mathrm{chg}}$ & Charging C-rate\\
7 & $\log (\mathrm{var}(\Delta dQ/dV_{\mathrm{w3}-\mathrm{w0}}^{3.0V-3.6V}(V))$ & Variance of low-voltage incremental capacity feature (see Sec. \ref{sec:ica_feature_extraction})\\
8 & $\Delta Q_{\mathrm{w3}-\mathrm{w0}}^{3}$ & Change in DVA-based capacity $Q^{\mathrm{DVA,3}}$ (see Supplementary Information)\\
9 & $\log (\left|\mathrm{mean}(\Delta dQ/dV_{\mathrm{w3}-\mathrm{w0}}^{3.0V-3.6V}(V)\right|)$ & Mean of low-voltage incremental capacity feature (see Sec. \ref{sec:ica_feature_extraction})\\
10 & $\log (\left|\mathrm{mean}(\Delta Q_{\mathrm{w3}-\mathrm{w0}}(V)\right|)$ & Mean of $\Delta Q(V)$ vector (see Sec. \ref{sec:feature_extraction})\\
\bottomrule
\end{tabular}
\end{adjustbox}
\end{table}
\renewcommand{\arraystretch}{1}

\subsection{Predicting Lifetime Using Machine Learning Models}

To predict the lifetime, we initially establish a pair of baseline models. 
The first baseline model is a dummy model that does not use any input features or have any trainable parameters, and instead predicts the mean cell lifetime of the training set for all cells. This is a good way to determine if a more complex model is truly learning new information from the input data, or instead only appears to be learning because of similar train/test dataset distributions that lead to similar error metrics. When tested on the two test datasets, the dummy model achieves $\mathrm{MAPE_{EOL}}$ of 31.52\% and 47.54\% on the high-DoD and low-DoD test sets, respectively. The error metrics for all models tested are shown in Table \ref{table:main_results}. 

 \renewcommand{\arraystretch}{1.5}
\begin{table}
\caption{Prediction errors for selected models tested using the high- and low-DoD test datasets.}
\centering
\begin{adjustbox}{max width=\textwidth}
\begin{tabular}{lccccccc}
\hline
\textbf{Model} & \textbf{$\mathbf{\mathit{N}}$ Features} & \multicolumn{3}{c}{\textbf{MAPE [\%]}} & \multicolumn{3}{c}{\textbf{RMSE [weeks]}} \\ & & Training
                              & High DoD        & Low DoD  
                              &Training
                              & High DoD          & Low DoD         \\ \hline
Dummy Model     & 0          & 35.0          & 31.5    & 47.5        &6.5    & 4.8             & 18.5             \\
Cycling Conditions   & 3       & 24.8  & 19.0          & 23.7           &4.0  & 3.3             & 9.8             \\
Discharge Model \citep{severson2019data}   & 5$^*$       &23.9  & 28.0          & 24.8        & 4.6   & 4.7             & 11.5             \\
Degradation-informed  & 2 & 17.3    & 16.0    &24.4   &3.2   &3.0   &7.8\\
Degradation-informed & 3 & 16.5 & 15.1 & 33.0          & 3.1            & 2.8             & 9.7              \\
HBM  & 2$^\dagger$  & 18.6   & 16.9    & 21.8   & 3.3  & 3.1      &7.3\\
HBM  & 3$^\dagger$  & 17.4   & 15.8    & 24.1   & 3.1  & 2.9      &7.5\\ \hline
\end{tabular}
\end{adjustbox}

{\raggedright $^{*}$ The discharge model \citep{severson2019data} contains six features, with one of them being the difference between the maximum capacity and capacity at cycle two, $\Delta Q_{\mathrm{max}-2}$. However, this feature cannot be calculated for our dataset due to the partial depth of discharge cycling and the continuously decreasing capacity-fade curves for all cells and has thus been omitted. 

$^{\dagger}$ The number of features listed refers to the number of cell-level input features. For both HBMs, a single cycling condition-level feature is used for grouping cells, and, as indicated in the table, either two or three cell-level features are used for regression.
\par}
\label{table:main_results}
\end{table}
\renewcommand{\arraystretch}{1}

The second baseline model is built using only the cycling condition parameters as input features. This model predicts lifetimes without using cell-specific aging measurements. This model achieves a $\mathrm{MAPE_{EOL}}$ of 19.01\% and 23.72\% on the high DoD and low DoD test sets, respectively. The substantial decrease in prediction error over the dummy model shows that the usage parameters convey a significant amount of information that can be used to predict lifetime accurately. This result is expected, as a great deal of battery lifetime modeling work \citep{smith2021lithium, gasper2022machine, jiang2021bayesian} has already explored the strong connection between usage and degradation. However, only using condition-level cycling features does not account for intrinsic cell-to-cell variability. Hence, the next set of models we tested included cell-level features extracted from the early aging data.

The first cell-level features model is the ``discharge model'' described in \citep{severson2019data} and Section \ref{sec:feature_extraction}. This model, and all other models built on cell-level inputs, use features extracted from the RPTs of weeks zero and three, which is just under 18\% of the average lifetime. The main feature included is $\mathrm{var}(\Delta Q_{\mathrm{w3}-\mathrm{w0}}(V))$, however, we found that this did not completely describe the variance in our dataset. When tested on the high and low DoD test datasets, the discharge model achieved 28.03\% and 24.80\% $\mathrm{MAPE_{EOL}}$, respectively. The performance on the two test datasets is slightly worse than the cycling condition model, yet still better than the dummy model, indicating that the features used in the discharge model do carry useful information, but are not optimal for our dataset (see Table \ref{table:main_results}). 

The remaining models we compare are the degradation-informed and hierarchical Bayesian models. We refer to our elastic net models as \emph{degradation-informed} in Table \ref{table:main_results} because of the newly developed degradation-based features used as model inputs. Both the degradation-informed and HBM models use the same sets of input features, and for thoroughness, we compare models built using two and three features each. Compared to the cycling condition baseline, the two-feature elastic net model shows decreased $\mathrm{MAPE_{EOL}}$ on the high-DoD test of 16.0\% and a slight increase in error on the low-DoD test set to 24.4\%. However, the $\mathrm{RMSE_{EOL}}$ of the low-DoD test set drops considerably from 9.8 to 7.8 weeks. For the HBMs, we observe small increases in the training and the high-DoD test errors while a noticeable improvement in the low-DoD test errors over the degradation-informed models using the same set of features.

For both the degradation-informed and hierarchical models, we observe that including the third feature decreases model prediction error on the training and high-DoD test datasets but increases error for the low-DoD test dataset. When the third feature is added, both models over-fit the training dataset and exhibit poor extrapolation capability to the low-DoD test dataset where the cells have longer lifetimes. Regardless, the HBM trained with three features still performs better when predicting the low-DoD test set compared with its elastic net counterpart. Generally, by comparing the evaluation metrics of the two models (degradation-informed model and HBM), we find that the HBM has better generalizability to the low-DoD test set, but at the cost of slightly higher training and high-DoD test errors.

The large improvement in performance observed for models using cell-level (as opposed to only using cycling condition features) features prompts us to further investigate why the feature $\log (\mathrm{mean}(\Delta dQ/dV_{\mathrm{w3}-\mathrm{w0}}^{3.6V-3.9V}(V))$ explains cell-to-cell variability better than other features. Firstly, it is more accurate to use measured health metrics from individual cells in operation to predict their lifetime. This reveals the intrinsic cell-to-cell variability that could cause different aging behaviors under identical cycling conditions. Secondly, this optimized feature, which likely captures how much loss of active material happens during early life, has a balanced representation of the variability within the group and among the entire dataset.

In summary, we find that the best feature $\log (\mathrm{mean}(\Delta dQ/dV_{\mathrm{w3}-\mathrm{w0}}^{3.6V-3.9V}(V))$ explains the cell-to-cell variability well for a majority of cells.  
The remaining variance in the feature-lifetime correlation may be contributed jointly by measurement inaccuracy and unexplained manufacturing variability. Hence, our analysis of the results suggests that a predictive early-life feature should capture the variability introduced by the difference in cycling conditions and information about intrinsic cell-to-cell variation that causes different performances under identical loads. Also, our feature engineering methodology (Sec.\ \ref{sec:ica_feature_extraction}) can be extended to find good features for other cell chemistries.

\begin{figure}
	\includegraphics[width=1\textwidth]{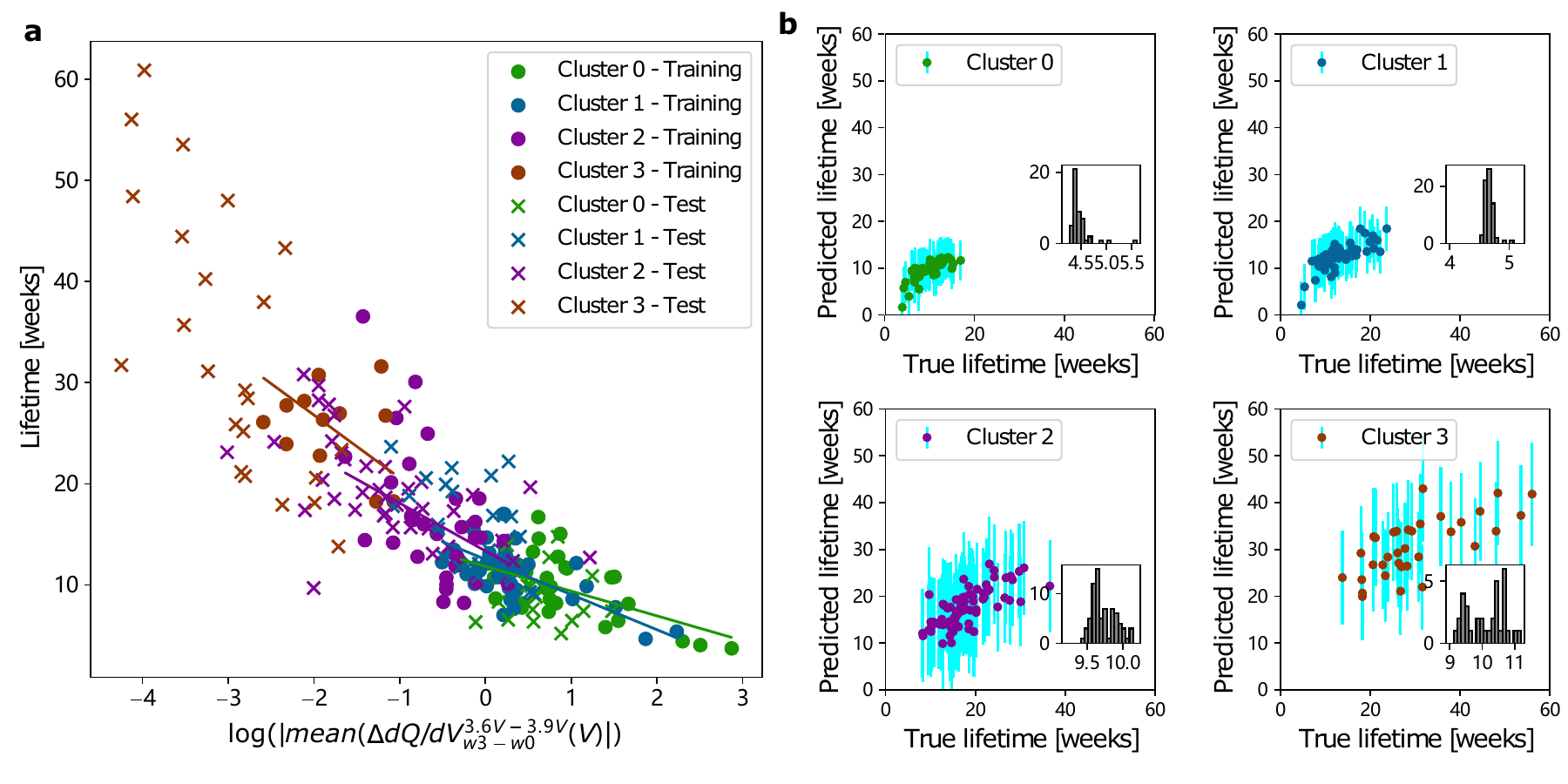}
	\centering
	\caption{Overview of HBM results. \textbf{a}, Relationship between $\log (\left|\text{mean}(\Delta dQ/dV_{\mathrm{w3}-\mathrm{w0}}^{3.6V-3.9V}(V)\right|)$ and true lifetime across different clusters and train-test split ("Test" denotes samples from both high- and low-DoD sets). Fits, corresponding to mean parameter values, are plotted for each cluster. \textbf{b}, Predictions for each cluster with 2 standard deviations as the corresponding error bar for each sample. The embedded histograms show a summary of error bars}
	\label{fig:hbm_joule_results}
 \end{figure}
 
\subsubsection{Analysis of HBM Results}
The probabilistic nature of HBMs enables us to extract a deeper understanding by considering both the mean and the uncertainty of lifetime predictions.  Assuming individual cluster fitting parameters and noise variance, $\bm{\theta_j}$ and $\sigma_j$ respectively, are independent, the posterior predictive distribution can be written as
\begin{equation}
p\left(y_j^* \mid Y_j\right)=\iint p\left(\sigma_j \mid Y_j\right) p\left(\bm{\theta_j} \mid Y_j\right) p\left(y_j^* \mid \bm{\theta_j}, \sigma_j\right) d \bm{\theta_j} d \sigma_j .
\end{equation}

For a point-wise prediction, one can estimate the mean value of $p\left(y_j^* \mid Y_j\right)$. Table \ref{table:main_results} lists the performance of the HBM built using two different feature sets. The first uses two cell-level features, $\log (\left|\mathrm{mean}(\Delta dQ/dV_{\mathrm{w3}-\mathrm{w0}}^{3.6V-3.9V}(V)\right|)$ and $\log (\Delta \mathrm{CV\; Time}_{\mathrm{w3}-\mathrm{w0}})$, and achieves 3.08 weeks $\mathrm{RMSE}$ and 16.88\% $\mathrm{MAPE}$ for the high-DoD test set, which is almost the same as the performance of the degradation-informed model using the same feature set. While, for the low-DoD test set, the HBM achieves 7.3 weeks $\mathrm{RMSE}$ and 21.83\% $\mathrm{MAPE}$, which outperforms the degradation-informed model by 7\% and 10\% for $\mathrm{RMSE}$ and $\mathrm{MAPE}$, respectively. 

Similar to the degradation-informed model, we observe that the HBM model overfits the training dataset when the third feature ($\mathrm{DoD}$) is added. This is evident by the increased performance on the training and high-DoD test set but worse performance on the low-DoD test set. Specifically, under the high-DoD test set, $\mathrm{RMSE}$ improved from 3.08 to 2.85 weeks, and $\mathrm{MAPE}$ improved from 16.88\% to 15.80\%. However, for the low-DoD test set, $\mathrm{RMSE}$ increased from 7.30 to 7.49 weeks, and $\mathrm{MAPE}$ increased from 21.83\% to 24.10\%. Notably, the HBM shows more resistance to overfitting than the degradation-informed model, whose performance decreased substantially more than the HBM when the third feature was included in the feature set.

Fig.\ \ref{fig:hbm_joule_results}b shows the uncertainty (2 standard deviations) of $p\left(y_j^* \mid Y_j\right)$ for posterior lifetime predictions of each cluster. The uncertainty levels for clusters 0 and 1 are around $\pm$4.5 weeks (at 2 s.d.), whereas for clusters 2 and 3, the uncertainty levels are around $\pm$9.5 and 10.5 weeks, respectively, which reflects the model's uncertainty when predicting cells from unseen cycling conditions. According to Table \ref{table:train-test-split}, there are only 12 cells from cluster 3 in the training set, while there are 23 cells from cluster 3 in the Low-DoD test set. Due to the lack of data, the uncertainty for all regression parameters ($\bm{\theta}_3, \sigma_3$) for cluster 3 is much larger than that of clusters 0 and 1. On the other hand, as the prediction uncertainty becomes large for long-life cells, uncertainty itself can be used as an indicator to denote whether one should include more early-life data for feature calculation. 
For example, when running HBM in a forward mode (using the trained model to give predictions), for test samples in Cluster 3, large prediction uncertainty is observed (>10 weeks). One may consider including the 4th or 5th week of training data to retrain the model so that the prediction uncertainty on Cluster 3 test samples can be reduced. Since the used three weeks of training data only take up $7\%$ of the average lifetime for Cluster 3 samples, using 1-2 more weeks train data still only covers the very early stage of these long-life cells.

\renewcommand{\arraystretch}{1.5}
\begin{table}
\caption{Summary of train-test split for each cluster}
\centering
\begin{tabular}{lccccc}
\hline
\multicolumn{1}{l}{\textbf{Cluster ID}} &  &  & $\mathbf{\mathit{N}}$ \textbf{Samples} &  &  \\
 & \multicolumn{1}{l}{$\mathrm{Stress_{\mathrm{avg}}}$} & \multicolumn{1}{l}{Training} & \multicolumn{1}{l}{High-DoD test} & \multicolumn{1}{l}{Low-DoD test} &  \\ \hline
0   & 2.2 & 30   & 18    & 0   &  \\
1   & 1.9 & 41   & 24    & 4   &  \\
2   & 1.5 & 33   & 18    & 22  &  \\
3   & 1.0 & 12   & 0     & 23  &  \\ \hline
Total & 1.7   & 116  & 60    & 49  &  \\ \hline
\end{tabular}
\label{table:train-test-split}
\end{table}
\renewcommand{\arraystretch}{1}

Due to the page limits, further analysis of uncertainty for model parameters can be found in Supplymentary. This uncertainty on both lifetime predictions and model parameters can be more beneficial to real-world applications compared to only a point-wise prediction. For example, instead of knowing the exact EoL lifetime, customers care more about a warranty for the worst-case lifetime, which can be satisfied by using the standard deviation of prediction distributions.

\subsection{Conclusion}

In this study, we have developed two data-driven models to tackle the problem of battery early lifetime prediction on a large and unique aging dataset, which consists of 225 NMC cells cycled under a wide range of charge and discharge C-rates (0.5-\SI{3}{C}) and DoDs (4-100\%). Our feature engineering process identifies a new predictive feature, $\mathrm{mean}(\Delta dQ/dV_{\mathrm{w3}-\mathrm{w0}}^{3.60\mathrm{V}-3.90\mathrm{V}}(V))$, derived from incremental capacity curves and closely related to the degradation induced by loss of active materials. Also, our analysis shows that the widely used $\Delta Q(V)$ features in the existing early prediction literature may not explain cell-to-cell lifetime variability within our dataset. 

In terms of results, two distinct machine learning models are trained to predict the lifetime. Our degradation-informed model, trained using elastic net regression, yields 3.0 and 7.8 weeks $\mathrm{RMSE}$ and 15.1\% and 33.0\% $\mathrm{MAPE}$ on the high- and low-DoD test sets, respectively. The HBM produces 3.1 and 7.3 weeks $\mathrm{RMSE}$ and 16.9\% and 21.8\% $\mathrm{MAPE}$ for the high- and low-DoD test sets, respectively. While the HBM shows performance improvement for point-wise predictions on the low-DoD test set, it also gives uncertainty information for its predictions, which can be used in applications like the cell lifetime warranty. And we found that the uncertainty grows across groups with the decrease of cycling stress factor $\mathrm{Stress_{\mathrm{avg}}}$, which indicates the lack of observability for cell-to-cell differences from early-life features, and thus more cycling time range may need to be included for cells under mild cycling conditions.

A limitation of this work is that the models are demonstrated on battery aging data collected in a well-controlled laboratory setting under constant cycling conditions over the life of the cells. However, depending on the applications, battery data from real-world applications may be more variable and noisy, posing a challenge to feature extraction and lifetime prediction. To investigate this further, we will expand the dataset by aging cells using simulated electric grid duty cycles (e.g., simulating peak shaving and frequency regulation cycles). 

\section{EXPERIMENTAL PROCEDURES}
\label{sec:experimental_procedures}
\subsection{Data and CodeAvailability}

The battery aging dataset collected and used for this work is available for download at: \href{https://doi.org/10.25380/iastate.22582234}{https://doi.org/10.25380/iastate.22582234}. Please refer to the dataset as the ISU \& ILCC NMC/Gr battery aging dataset.

The code for the data preprocessing, feature extraction, and early prediction modeling is available at: \href{https://doi.org/10.25380/iastate.22582234}{https://doi.org/10.25380/iastate.22582234}.

\subsection{Cell and Tester Specifications}

The Li-ion cells used in this study were commercial 502030 size Li-polymer cells with nickel-manganese-cobalt (NMC) as the positive electrode and graphite as the negative electrode, manufactured by Honghaosheng Electronics in Shenzhen, China. The rated capacity is \SI{250}{mAh} (giving 1C as \SI{250}{mA}), and the operating voltage ranges from 3.0 to \SI{4.2}{V}. All cells were tested on two 64-channel Neware BTS4000 battery testers, in thermal chambers set at \SI{30}{\celsius}.

\subsection{Battery Aging Test Design}
\label{sec:test_design}
The aging experiments were designed around three main stress factors that impact battery lifetime: charge rate ($\mathrm{C}_{\mathrm{chg}}$), discharge rate ($\mathrm{C}_{\mathrm{dis}}$), and depth of discharge ($\mathrm{DoD}$). To track the full discharge capacity of cells with partial depths of discharge cycling, we periodically ran RPTs that measured cell capacity and gathered complete $Q(V)$ data for feature engineering. Each RPT consisted of two cycles performed at slow rates (C/2 and C/5) to capture cell voltage response while minimizing the impact of the cell kinetics. Before beginning the aging tests, an initial RPT was conducted to determine the beginning-of-life health. Aging tests consisted of 1 week of cycling followed by an RPT, and they were repeated until cell capacity decreased below \SI{200}{mAh} (80\% of the rated capacity). %
%

As previously mentioned, four cells were cycled at each test condition. We refer to a specific cell using its group number and cell identifier, e.g., G7C3, where the numbers following each letter indicate the group and cell, respectively. Initially, we aimed to study two stress factors: $\mathrm{DoD}$ and $\mathrm{C}_{\mathrm{chg}}$. Conditions were selected using a grid search, with the discharge rate fixed at 0.5C for all cells. Later, we expanded the dataset to study the third stress factor, $\mathrm{C}_{\mathrm{dis}}$. Additional conditions were then selected using random sampling. The charge/discharge rates and depths of discharge were sampled evenly from the ranges 0.5C to 3C and 25\% to 100\%, respectively.

The cycling conditions for all cell groups can be found in Supplementary Information Table S1. However, the depth of discharge design values do not exactly match the measured depths of discharge from the cycling experiments. When we programmed the cycling protocols, we determined the cutoff voltages using a reference discharge capacity vs.\ voltage curve from a cell cycled at C/2. Unfortunately, the voltage hysteresis that the cells experience under C/2 discharge causes the cells to reach the cutoff voltage quicker than expected, thus causing the difference between the measured and designed depth of discharge. In this paper, we present and discuss the depth of discharge using the actual measured values since they more accurately represent the test conditions the cells experienced.

\subsection{Extracting Features from Incremental Capacity Data}
\label{sec:ica_feature_extraction}

Extracting features from incremental capacity curves is a natural extension to using the $Q(V)$ discharge curve since it is defined over the same fixed voltage range for every cell. After fitting a spline and downsampling each cell’s $Q(V)$ curve to 1000 points, we calculated incremental capacity $(dQ/dV(V))$ as a finite difference approximation (difference quotient) of the first derivative of $Q(V)$ based on measurements of the $Q$ and $V$ time series \citep{severson2019data}. It is well documented that incremental capacity analysis is an effective method for cell degradation diagnostics \citep{birkl2017degradation, pastor2017comparison, dubarry2022best}. Measuring changes to the incremental capacity curve over the lifetime enables the diagnosis of different degradation modes, specifically loss of lithium inventory, and loss of active material in each electrode. Hence, we calculate core summary statistics of $\Delta dQ/dV(V)$ over a partial voltage range so as to focus the feature extraction on specific areas that may correspond to specific degradation modes. This approach is inspired by work in \citep{greenbank2021automated}, where the authors showed a strong correlation between the time a cell spends in a specific voltage range and its capacity loss, although here the incremental capacity curve is a result of degradation rather than a cause. Instead of manually specifying the voltage range to calculate the summary statistics, we exhaustively searched the entire 3.0 to \SI{4.2}{V} range in increments of \SI{0.01}{V}, with a minimum window size of \SI{0.02}{V} searching for the maximum Pearson correlation coefficient. 

\subsection{Extracting Features from Cycling Conditions}
\label{sec:stress_features}

As briefly mentioned in Sec. \ref{fig:f_eng}, we consider a set of stress-related features for early life prediction, which is $\mathrm{Stress_{\mathrm{chg}}}={\mathrm{C}_{\mathrm{chg}}}^{0.5} \mathrm{DoD}^{0.5}$. This feature captures the square-root-of-cycling charge throughput and is a proxy for diffusion-induced stress in the electrode active materials \citep{smith2021lithium, reniers2019review, smith2006solid}. In addition to the charge-based feature, we also calculate a discharge feature, $\mathrm{Stress_{\mathrm{dchg}}}={\mathrm{C}_{\mathrm{dchg}}}^{0.5} \mathrm{DoD}^{0.5}$. Further, to capture the effects of different charge and discharge rates in a single feature, we calculate an average stress feature as $\mathrm{Stress_{\mathrm{avg}}} = (\mathrm{Stress_{\mathrm{chg}}} + \mathrm{Stress_{\mathrm{dchg}}})/2$ and also calculate a multiplicative stress feature as $\mathrm{Stress_{mult}}=\mathrm{Stress_{\mathrm{chg}}} \cdot \mathrm{Stress_{\mathrm{dchg}}}$. For all features, we use the measured DoD from the first week of cycling in the calculation. A unique characteristic of these features is that they require no cell-specific measurements, assuming the calculation of DoD is accurate and accounts for voltage hysteresis. For this reason, these features are excellent candidates as condition-level grouping variables in our hierarchical Bayesian modeling approach to early prediction (see Sec. \ref{sec:hbm_for_early_prediction}).

\subsection{Feature Selection}
\label{sec:feature_selection}

We have so far focused on features that quantify the rate of degradation and correlate strongly with lifetime. However, simply using all the extracted features as inputs to a machine learning model may yield poor results for two reasons. First, some features are strongly correlated with each other, known as multicollinearity. A model trained with collinear features can be sensitive to minor changes in the feature values and may extrapolate poorly  \citep{dormann2013collinearity}. Second, while our dataset is large compared to existing publicly available datasets (225 cells), it is still relatively small from a machine learning perspective. Small datasets require special care to avoid over-fitting and improve generalization performance on unseen test data. This is especially the case when the number of data points is not significantly larger than the number of features $(N_{\mathrm{data}} \gg N_{\mathrm{features}})$. Therefore, it is crucial to select a subset of highly predictive features before model training   \citep{cai2018fea_select,element_stat_learn}. 

To reduce the number of input features, we perform step-wise forward selection using a linear model and repeated cross-validation with $\mathrm{RMSE_{EOL}}$ as the evaluation metric. Starting with a null model, one feature is added to the model for each step until the number of selected features reaches a preset threshold ($N=10$). During each step, all features are tested in the model, and the feature that reduces the mean of the cross-validation $\mathrm{RMSE_{EOL}}$ the most is selected and added to the model for the next step. Simultaneously, we evaluated the selected model at each step using the standard deviation of the cross-validation $\mathrm{RMSE_{EOL}}$. We then select the features to use corresponding to the set with a balance between low mean and small standard deviation of cross-validated $\mathrm{RMSE_{EOL}}$. In practice, we tend toward selecting fewer features so that the resulting model will be less complex and extrapolate better.

\subsection{Elastic Net Regression for Lifetime Prediction}
To predict cell lifetime, we formulate a regression problem with the extracted early-life features $\mathbf{X}=\left[\mathbf{x}_1,\mathbf{x}_2,...,\mathbf{x}_m\right]$ as inputs, and the measured cell lifetimes $\mathbf{y}=\left[y_1,y_2,...,y_n\right]^T$ in logarithmic scale as outputs, where $m$ is the number of early-life features, and $n$ is the number of cells. Each element of $\mathbf{X}$ is a column vector containing the specific features selected through the technique introduced in Sec.\ \ref{sec:feature_selection}. We assume that the lifetime is a linear function of the early-life features, giving 
\begin{equation}
\hat{y}=f(\mathbf{X})=\boldsymbol{\beta}_0+\mathbf{X}\boldsymbol{\beta}_1,
\end{equation}
where $\boldsymbol{\beta}_0$ is an $n\times1$ column vector of intercepts and $\boldsymbol{\beta}_1$ is a vector of coefficients, one for each feature, $\boldsymbol{\beta}_1=\left[ \beta_1, \beta_2, ...,\beta_m\right]^T$. 

To find the coefficients of this equation, we formulate an optimization problem with elastic net regularization, which is a combination of $\mathrm{L}_1$ and $\mathrm{L}_2$ penalization. The objective function is
\begin{equation}
    \hat{\boldsymbol{\beta}} = \underset{\beta_0, \boldsymbol{\beta}_1}{\operatorname{argmin}}\left( \|\mathbf{y}-\boldsymbol{\beta}_0-\mathbf{X}\boldsymbol{\beta}_1)\|_2^2 +\lambda\left(\frac{1-\alpha}{2}\|\boldsymbol{\beta}\|_2^2+\alpha \|\boldsymbol{\beta}\|_1\right) \right),
\end{equation}
where $\alpha$ and $\lambda$ are hyperparameters that control the balance between the $\mathrm{L}_1$ and $\mathrm{L}_2$ penalties and the magnitude of regularization, respectively. To select optimal values of $\alpha$ and $\lambda$, we perform repeated cross-validation using randomized dataset splits.

\subsection{Hierarchical Bayesian Models for Early Prediction}
\label{sec:hbm_for_early_prediction}

As a comparison and contrast to the method in the previous section, we also consider hierarchical Bayesian models (HBMs) for lifetime prediction. These have a layered structure that can model changes in the feature-target relationship throughout the dataset. HBMs have been applied to model naturally structured data in various research fields from ecology to sociology, psychology, and computer vision \citep{lake2015human,pedersen2019hierarchical}. 

\subsubsection{Clustering for Hierarchical Modeling}

For our problem of early life prediction, features can be viewed as coming from two levels: the `cycling condition' level and the `individual cell' level. Condition-level features relate to user-defined test protocols rather than measured data.
For our dataset, the charge/discharge C-rates and depth of discharge ($\mathrm{C}_{\mathrm{chg}}$, $\mathrm{C}_{\mathrm{dchg}}$, $\mathrm{DoD}$), and any mathematical combination of these are all condition-level features. In contrast, features that require specific cell measurements during cycling are considered cell-level features. Features such as $\mathrm{mean}\left(\Delta dQ/dV_{\mathrm{w3}-\mathrm{w0}}^{3.60\mathrm{V}-3.90\mathrm{V}}(V)\right)$ and $\mathrm{var}\left(\Delta Q_{\mathrm{w3}-\mathrm{w0}}(V)\right)$ are examples of cell-level features that are unique to each cell. 

To validate the hypothesis that conditional-level features have a strong impact on the relationship between cell-level features and lifetime, we calculate the condition-level feature $\mathrm{Stress_{\mathrm{avg}}} = ({\mathrm{C}_{\mathrm{chg}}}^{0.5} \mathrm{DoD}^{0.5} + {\mathrm{C}_{\mathrm{dchg}}}^{0.5} \mathrm{DoD}^{0.5})/2$ described in Sec.\ \ref{sec:stress_features}. This represents the average diffusion-induced stress that a cell experiences \citep{smith2021lithium}.
To take advantage of an HBM's ability to model the change in feature-target relationship across different levels, we investigate clustering cell data based on cycling conditions, quantified by average stress ($\mathrm{Stress_{\mathrm{avg}}}$). In general, we expect cells with similar average stress levels to share the same feature-lifetime relationship, enabling the HBM to better fit the dataset. We adopt a constrained K-means clustering algorithm \citep{bhattacharya2018faster}, which is an improved version of the traditional K-means algorithm that imposes minimum and maximum cluster size limits. A optimal cluster number 
 $K=4$ is used in later analysis, details can be found in Supplementary Information.

\subsubsection{Bayesian Hierarchical Linear Model}
\pgfdeclarelayer{background}
\pgfdeclarelayer{foreground}
\pgfsetlayers{background,main,foreground}

\definecolor{bbb}{rgb}{.679, .835, .941}
\definecolor{rrr}{rgb}{.66, .871, .746}
\tikzstyle{sensor}=[draw, fill=bbb, text width=9em, 
    text centered, minimum height=4em, rounded corners]
\tikzstyle{ann} = [above, text width=5em]
\tikzstyle{naveqs} = [sensor, text width=9em, fill=rrr, 
    minimum height=6em, rounded corners]

\begin{figure*}
\centering

\begin{tikzpicture}
    \node (hypers) [naveqs] {$\begin{aligned} & \gamma \sim N(\bar{\gamma}, 10) \\ & \sigma \sim \operatorname{HalfCauchy}(1)\end{aligned}$};
    \path (hypers.140)+(+6.5,0) node (theta0) [sensor] {$\begin{aligned} & \theta_0=\bm{\gamma}^{\top} g_0 \\ & \sigma_0 \sim \operatorname{HalfCauchy}(\sigma)\end{aligned}$};
    \path (hypers.0)+(+1.5,0) node (mid) {\dots};
    \path (hypers.0)+(+6.5,0) node (mid) {\dots};
    
    \path (hypers.-140)+(+6.5,0) node (thetaj) [sensor] {$\begin{aligned} & \theta_J=\bm{\gamma}^{\top} g_J \\ & \sigma_J \sim \operatorname{HalfCauchy}(\sigma)\end{aligned}$};

    \path (hypers.140)+(+11.5,0) node (y0i) [sensor] {$\begin{aligned} & y_{0 i}=\theta_0^{\top} x_{0 i}+\varepsilon_0 \\ & \varepsilon_0 \sim N\left(0, \sigma_0\right)\end{aligned}$};
    \path (hypers.-140)+(+11.5,0) node (yji) [sensor] {$\begin{aligned} & y_{J i}=\theta_J^{\top} x_{J i}+\varepsilon_J \\ & \varepsilon_J \sim N\left(0, \sigma_J\right)\end{aligned}$};

    \path (hypers.north)+(-0.0,+0.4) node (cyc) {Hyper priors};
    \path (theta0.north)+(-0.0,+0.4) node (param) {Cluster parameters};
    \path (y0i.north)+(-0.0,+0.4) node (pred) {Sample predictions};
    
    \path (hypers.south east)+(+0.5,-2.0) node (INS) {Condition-level regression};
    \path (hypers.south east)+(+5.5,-2.0) node (INS) {Cell-level regression};
    
    \draw [->] (hypers.0) -> (theta0.west);
    \draw [->] (hypers.0) -> (thetaj.west);

    \draw [->] (theta0.0) -> (y0i.west);
    \draw [->] (thetaj.0) -> (yji.west);
    
    \begin{pgfonlayer}{background}
        \path (theta0.north west)+(-0.5,1.1) node (a) {};
        \path (yji.south east)+(+0.3,-0.8) node (b) {};
        
        \path[fill=blue!10,rounded corners, draw=black!80, dashed]
            (a) rectangle (b);
        
        \path (cyc.north west)+(-1,0.8) node (a) {};
        \path (thetaj.south east)+(+0.5,-0.4) node (b) {};
        \path[fill=yellow!10,rounded corners, draw=black!80]
            (a) rectangle (b);

    \end{pgfonlayer}
\end{tikzpicture}
     \caption{Overview of HBM structure. Model parameters can be classified as either individual-level  ($\bm{\theta_{j}},\bm{\sigma_{j}}$) or conditional-level  ($\bm{\gamma},\bm{\sigma}$); $j$ represents cycling condition group index, $i$ represents individual cell index, $y_{ji}$ represents lifetime of $i$th cell in $j$th cycling group. The two-level structure allows the individual cell-level feature-label ($x_{ji}-y_{ji}$) relationship to vary with cycling condition based on cycling condition level features ($\bm{g_{j}}$).}
     \label{fig:modelstructure}
\end{figure*}
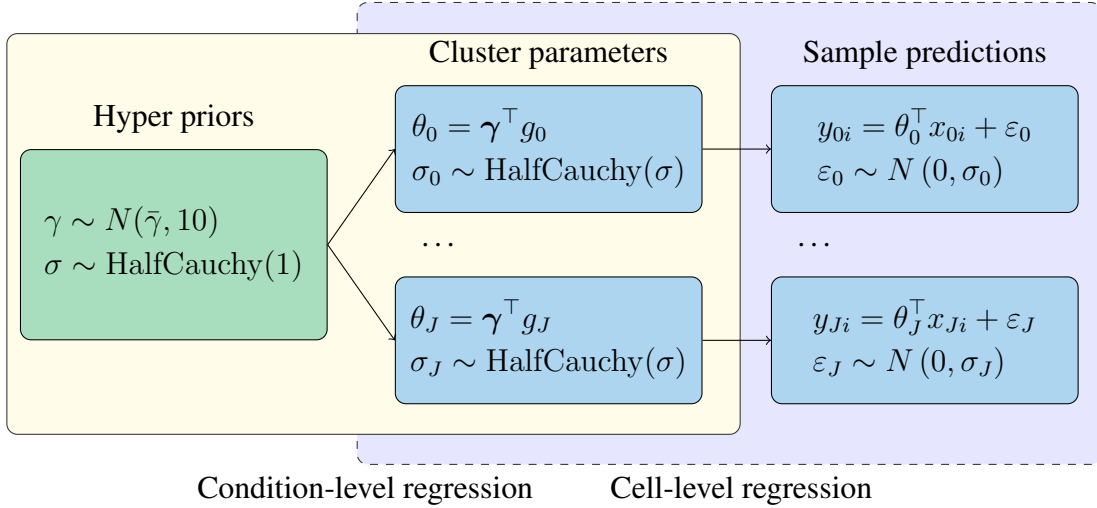

Similar to the HBM used in former work \citep{zhou2022bayesian}, our model structure has two levels and is shown in Fig.\ \ref{fig:modelstructure}. The first level considers the cycling condition parameters. As mentioned previously, cells are first divided into four clusters (indexed from 0) based on their average stress $\mathrm{Stress_{\mathrm{avg}}}$, calculated using the cycling condition parameters. %

At this level, we aim to find the mapping (parameterized by $\bm{\gamma}, \sigma$) between condition-level features ($\bm{g_{j}}$) and the cell-level regression parameters ($\bm{\theta_{j}},\sigma_j$). 
\begin{equation}
\begin{aligned} & \theta_j=\bm{\gamma}^{\top} \bm{g_j} \\ & \sigma_j \sim \operatorname{HalfCauchy}(\sigma)\end{aligned}
\end{equation}

After the coefficients ($\bm{\theta_{j}},\sigma_j$) are decided for each cluster, the individual cell-level regression is built as the second level of the HBM. The cell-level regression uses individual health features ($\bm{x_{ji}}$) and coefficients ($\bm{\theta_{j}},\sigma_j$) to give lifetime predictions ($\bm{y_{ji}}$) for individual cells.

\begin{equation}
    y_{j i} \sim N(\bm{\theta_{j}}^{\top} \bm{x_{j i}}, \sigma_{j}^2)
\end{equation}

The overall training objective is to infer posterior distributions for both the condition-level model and the individual cell-level models, $P\left(\bm{\theta_{j}} \mid Y_{j} \right)$ and $P\left(\bm{\gamma} \mid\left\{Y\right\}\right)$ respectively, where $Y_{j}$ represents lifetimes from only the $j$th group but $\left\{Y\right\}$ represents data from all lifetimes. More details about the training procedure and hyper-priors are included in Supplementary Information.

\subsection{Model Evaluation Metrics}
\label{sec:evaluation_metrics}

We use two standard error metrics to evaluate the lifetime prediction accuracy of our approaches, namely, mean absolute percentage error ($\mathrm{MAPE}_{\mathrm{EOL}}$) and root mean squared error ($\mathrm{RMSE}_{\mathrm{EOL}}$), both calculated using the measured and predicted values of cell lifetime on a linear scale. The metrics are

\begin{equation}
    \mathrm{MAPE}_{\mathrm{EOL}}=\frac{1}{n} \sum_{i=1}^n \left| \frac{\mathbf{y}_i-\hat{\mathbf{y}_i}}{\mathbf{y}_i} \right|\times 100\%
\end{equation}

\begin{equation}
    \mathrm{RMSE}_{\mathrm{EOL}}=\sqrt{\frac{1}{n}\sum_{i=1}^n(\mathbf{y}_i-\hat{\mathbf{y}_i})^2}
\end{equation}
where $\mathbf{y}$ are the measured cell lifetimes, $\hat{\mathbf{y}}$ are the predicted cell lifetimes, and $n$ is the number of cells.

\section*{AUTHOR CONTRIBUTIONS}

Conceptualization, T.L., A.T., Z.Z., C.H., D.H.; Data Collection, Data Management, Raw Data Processing, T.L.; Investigation, Methodology, Visualization, Software, Formal Analysis, Writing -- Original Draft, T.L., A.T., Z.Z.; Writing -- Review and Editing, T.L., A.T., Z.Z., C.H., D.H.

\section*{ACKNOWLEDGEMENTS}

We acknowledge the hard work of Jinqiang Liu from Iowa State University and Chad Tischer and Reuben D.\ Schooley from Iowa Lakes Community College for executing and maintaining the battery aging tests. We also want to acknowledge Murtaza Zohair for assembling the half-cells used in this study.

The work at Iowa State University and the University of Connecticut was partly supported by Iowa Economic Development Authority under the Iowa Energy Center Grant No.\ 20-IEC-018 and partly by the US National Science Foundation under Grant No.\ ECCS-2015710. The China Scholarship Council and the Department of Engineering Science supported the work at the University of Oxford. Any opinions, findings, or conclusions in this paper are those of the authors and do not necessarily reflect the sponsors' views.

\section*{DECLARATION OF INTERESTS}
The authors declare no competing interests. D.H. is a co-founder of Brill Power Ltd.

\newpage
\section*{SUPPLEMENTARY INFORMATION}
\subsection*{Detailed Battery Test Conditions}

After completing the aging tests on a small batch of cells, we found the capacity loss during the first week of cycling was extremely large. To better capture the capacity fade trend in this region during future tests, we added an extra RPT after 0.5 weeks of cycling. The supplementary material accompanying this paper specifies the group of cells with the extra RPT at week 0.5. This note is important since it affects feature extraction methods that use numerical indexing to select the appropriate data.

Starting from G20, we performed an additional RPT at week 0.5. By the time this manuscript was prepared, G49 and G57 hadn't had any cells reach the end of life. So, these two groups of cells were not included in this study, but the data will be available with other groups in this dataset. G15 is omitted from the dataset because of severe degradation within the first week of cycling and cells were removed for a safety concern.

\begin{longtblr}[
caption = {Cycling conditions of all groups},
label={table:supplementary-conditions}
]{ hlines,
vlines,
rowhead = 1,
}
\centering
Group \# & Charging C-rate & Discharging C-rate & Mean DoD    & Mean Lifetime [weeks]\\ 
1       & 0.500           & 0.500              & 25.9\% & 49.27    \\
2       & 0.500           & 0.500              & 46.1\% & 26.49    \\
3       & 0.500           & 0.500              & 67.8\% & 23.71    \\
4       & 1.000           & 0.500              & 47.3\% & 26.71    \\
5       & 1.000           & 0.500              & 97.5\% & 12.45    \\
6       & 2.000           & 0.500              & 27.2\% & 42.92    \\
7       & 2.000           & 0.500              & 46.9\% & 26.93    \\
8       & 2.000           & 0.500              & 67.2\% & 17.43    \\
9       & 2.000           & 0.500              & 96.9\% & 6.02     \\
10      & 2.500           & 0.500              & 47.2\% & 18.58    \\
12      & 3.000           & 0.500              & 28.6\% & 17.09    \\
13      & 3.000           & 0.500              & 47.3\% & 14.78    \\
14      & 3.000           & 0.500              & 66.3\% & 5.39     \\
16      & 0.500           & 0.500              & 97.6\% & 16.11    \\
17      & 1.000           & 0.500              & 67.5\% & 17.82    \\
18      & 2.500           & 0.500              & 27.5\% & 25.79    \\
19      & 2.500           & 0.500              & 65.7\% & 8.59     \\
20      & 0.800           & 0.500              & 98.2\% & 12.07    \\
21      & 1.200           & 0.500              & 98.0\% & 13.32    \\
22      & 1.400           & 0.500              & 97.9\% & 9.85     \\
23      & 1.600           & 0.500              & 97.6\% & 11.03    \\
24      & 1.800           & 0.500              & 97.5\% & 10.27    \\
25      & 1.800           & 0.600              & 51.4\% & 18.45    \\
26      & 1.400           & 2.200              & 4.2\%  & 39.79    \\
27      & 0.600           & 2.400              & 57.8\% & 15.06    \\
28      & 2.400           & 1.600              & 81.0\% & 9.03     \\
29      & 1.600           & 1.800              & 52.5\% & 15.05    \\
30      & 0.800           & 0.800              & 77.6\% & 13.31    \\
31      & 1.200           & 1.000              & 61.4\% & 11.83    \\
32      & 1.000           & 1.400              & 38.9\% & 18.72    \\
33      & 2.000           & 1.200              & 40.6\% & 17.77    \\
34      & 2.200           & 2.000              & 27.6\% & 21.40    \\
35      & 1.825           & 0.500              & 97.0\% & 10.67    \\
36      & 2.075           & 0.500              & 97.1\% & 10.23    \\
37      & 0.725           & 0.500              & 98.1\% & 15.32    \\
38      & 1.875           & 0.500              & 95.8\% & 10.62    \\
39      & 1.475           & 0.500              & 97.1\% & 11.53    \\
40      & 1.825           & 1.025              & 31.9\% & 28.97    \\
41      & 2.075           & 1.775              & 67.0\% & 14.33    \\
42      & 0.725           & 2.375              & 36.3\% & 18.67    \\
43      & 1.875           & 2.325              & 38.0\% & 20.21    \\
44      & 0.775           & 1.275              & 37.1\% & 19.25    \\
45      & 1.125           & 1.725              & 80.3\% & 8.44     \\
46      & 1.225           & 2.025              & 75.4\% & 8.84     \\
47      & 2.325           & 1.925              & 48.3\% & 11.81    \\
48      & 2.375           & 2.225              & 58.6\% & 9.66     \\
49      & 0.975           & 0.675              & 18.8\% &          \\
50      & 2.425           & 1.625              & 20.1\% & 20.77    \\
51      & 2.275           & 1.875              & 53.6\% & 11.80    \\
52      & 1.425           & 0.875              & 83.5\% & 15.63    \\
53      & 2.025           & 0.825              & 91.5\% & 5.72     \\
54      & 0.925           & 1.125              & 27.1\% & 27.19    \\
55      & 1.025           & 2.475              & 61.4\% & 12.48    \\
56      & 2.175           & 0.975              & 53.8\% & 18.46    \\
57      & 1.775           & 1.175              & 17.4\% &          \\
58      & 2.475           & 0.575              & 42.0\% & 19.36    \\
59      & 1.325           & 1.825              & 31.6\% & 25.78    \\
60      & 0.675           & 1.325              & 83.6\% & 10.67    \\
61      & 2.125           & 1.975              & 54.1\% & 11.67    \\
62      & 1.575           & 2.425              & 64.2\% & 11.54    \\
63      & 1.975           & 1.675              & 80.6\% & 10.12    \\
64      & 1.175           & 1.425              & 91.4\% & 8.16     \\ 

\end{longtblr}

\subsection*{Cycling Parameters vs. Lifetime}

\begin{figure}[h]
	\includegraphics[width=1\textwidth]{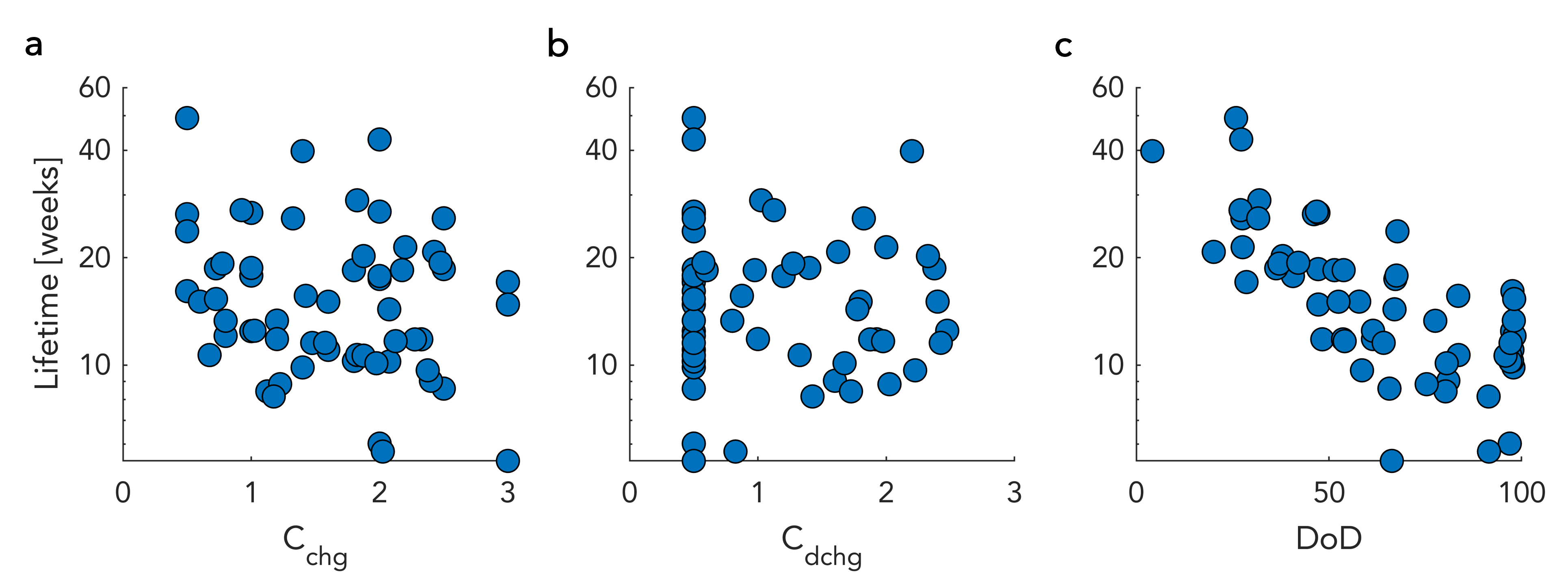}
	\centering
	\caption{Group Mean lifetime of  plotted as a function of \textbf{a}, $C_{\mathrm{chg}}$; \textbf{b}, $C_{\mathrm{dchg}}$; \textbf{c}, $DoD$. }
	\label{fig:SI_cond_lifetime}
	\centering
\end{figure}

\subsection*{Overview of Li-ion Battery Aging Under Group-Varying Conditions}
\label{sec:overview_capacity_fade}

\begin{figure}
	\includegraphics[width=0.99\textwidth]{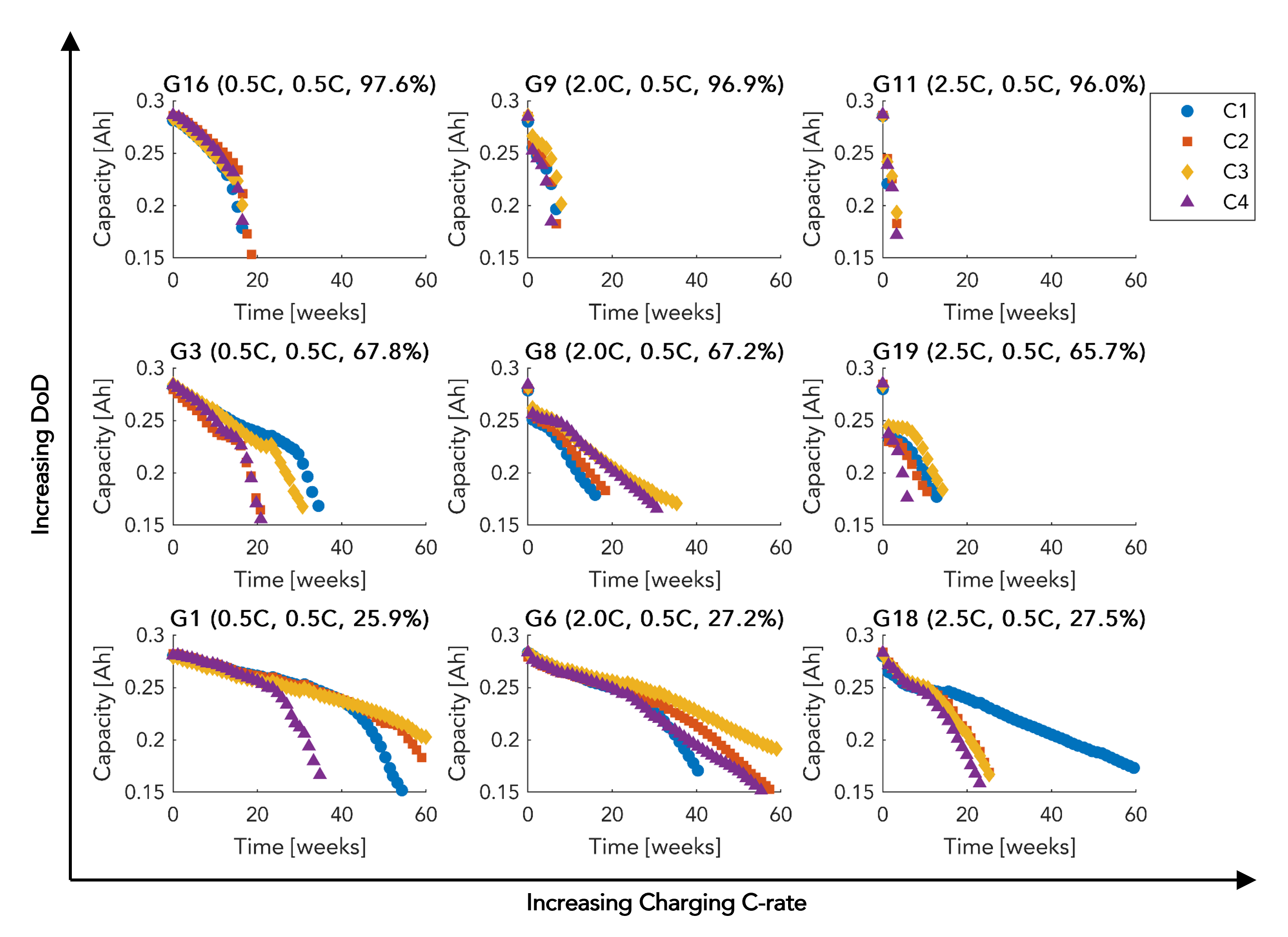}
	\centering
	\caption{Example capacity fade trajectories for groups cycled under different charging C-rates and DoDs. The values inside parentheses indicate charging C-rate, discharging C-rate, and mean DoD, respectively.}
	\label{fig:cap_fade_panel}
	\centering
\end{figure}

To showcase the many unique capacity-fade trajectories present in the dataset, we plot capacity-fade curves from groups of cells whose cycling conditions make up a complete grid spanning a range of charge C-rates and depths of discharge. This subset of 9 groups of cells, shown in Fig.\ \ref{fig:cap_fade_panel}, were cycled with different charging rates and depths of discharge but a constant discharge C-rate of 0.5C in all cases. We observe that groups with high charging rates and moderate-to-low depth of discharge (e.g., G8, G18, G19) experienced three-stage capacity fade. Their capacity initially decreases quickly, then stabilizes into a slower linear fade, and then accelerates again towards the end of life. More frequently, we observe a two-stage capacity fade trend from cells in some groups (e.g.\ G1, G3, G6, G16). However, in a few cases, we also observe a one-stage capacity trend for cells cycled at full depth of discharge and high charging C-rates (e.g., G9, G11). Our dataset's diverse capacity degradation trajectories make early-life feature engineering challenging because cells experiencing rapid capacity fade during the first few weeks of aging can sometimes end up having moderately long lifetimes. For example, G18 cells in Fig.\ \ref{fig:cap_fade_panel} show rapid capacity fade during the first few weeks of cycling but eventually had lifetimes greater than 20 weeks. On the other hand, G16 cells show much slower capacity fade during the first few weeks but have lifetimes of less than 20 weeks.

\begin{figure}
	\includegraphics[width=0.75\textwidth]{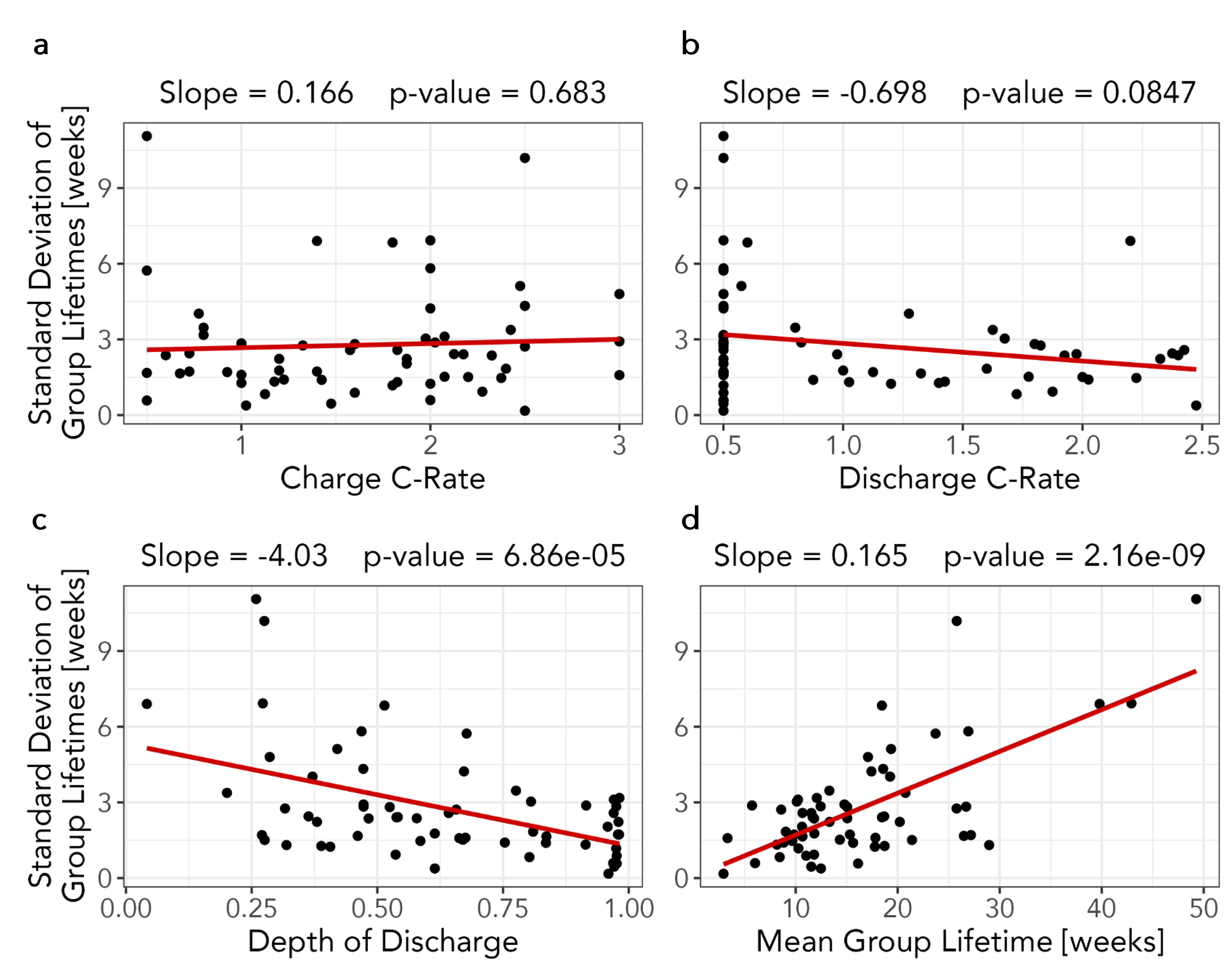}
	\centering
	\caption{Depth of discharge has a strong impact on lifetime variability. Here, the standard deviation of group lifetimes is plotted vs.\ \textbf{a,} charging C-rate, \textbf{b,} discharging C-rate, \textbf{c,} depth of discharge, \textbf{d,} the mean group lifetime. Smaller p-values indicate greater statistical significance of the fitted value of the slope term in the regression fit.}
	\label{fig:cond_group_life_variance}
	\centering
\end{figure}

Additionally, we observe considerable in-group lifetime variation. Groups G1, G6, and G18 in Fig.\ \ref{fig:cap_fade_panel} show a large variation in lifetime for cells operating under the same test conditions. Cell aging variability can be caused by testing equipment inaccuracies, manufacturing variations, and even internal defects, and is highly undesirable when designing battery-powered products. We conducted a statistical analysis to elucidate the relationship between the three cell-aging stress factors and lifetime variability. We calculated the in-group standard deviation of cell lifetime as a function of each aging stress factor, as well as the mean group lifetime, Fig.\ \ref{fig:cond_group_life_variance}. This reveals that only the depth of discharge (Fig.\ \ref{fig:cond_group_life_variance}c) has a statistically significant relationship with the observed cell-to-cell lifetime variability. However, we also observe that cells with longer lifetimes have higher lifetime variability (Fig.\ \ref{fig:cond_group_life_variance}d). These two results make it difficult to determine the true source of lifetime variability---it might result either from shallow depths of discharge or increased cell lifetime. 


\subsection*{Validation of $\Delta Q(V)$ Features}
\begin{figure}[h]
	\includegraphics[width=.5\textwidth]{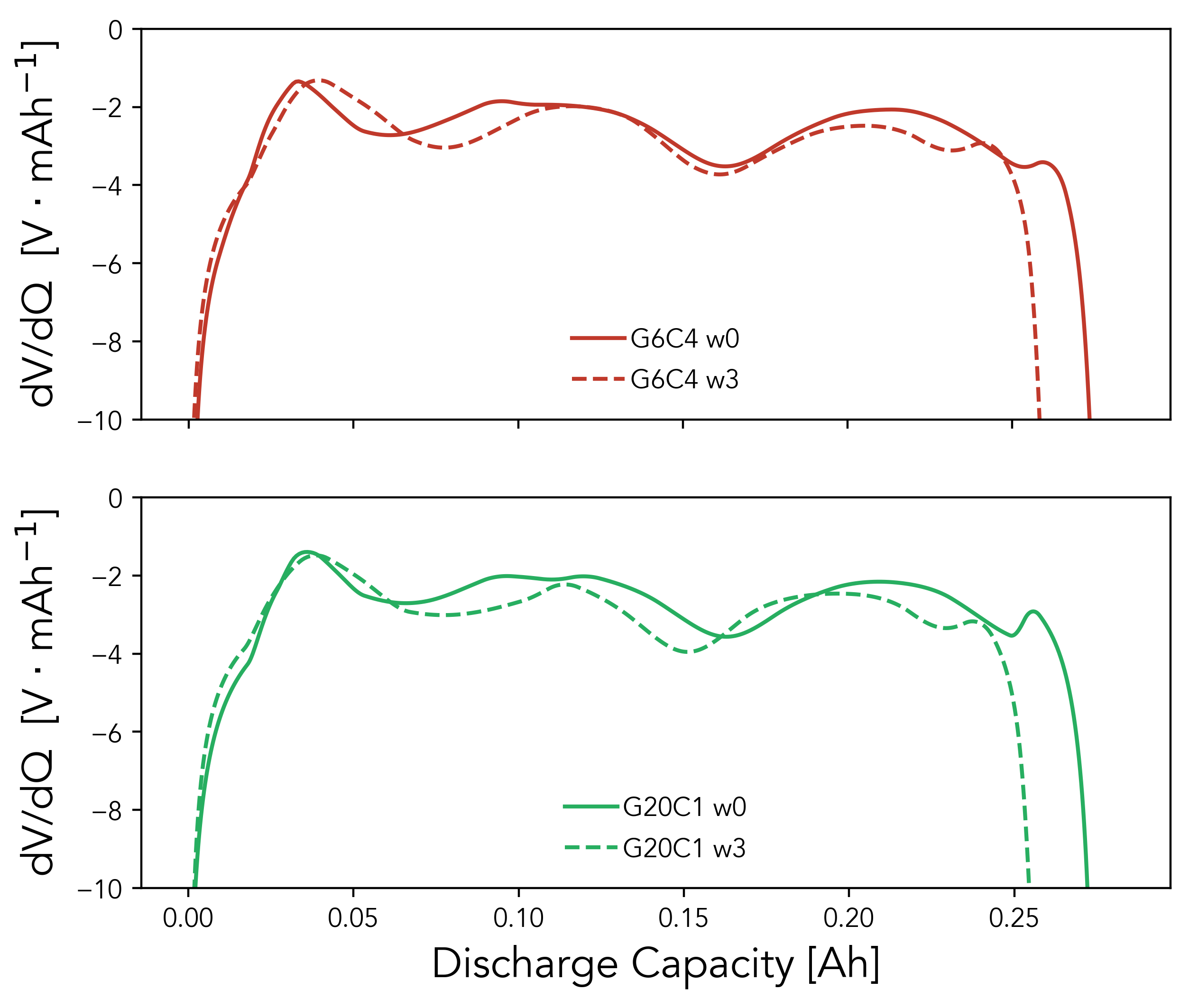}
	\centering
	\caption{$dQ/dV (V)$ curve changes between weeks 3 and 0 of two highlighted cells in the main text, which have similar $\mathrm{var}(\Delta Q(V)_\mathrm{w3-w0})$ values. }
	\label{fig:dQdV_check_deltaQ}
	\centering
 \end{figure}

Fig. \ref{fig:dQdV_check_deltaQ} shows the $dV/dQ (Q)$ curve evolution of two cells highlighted in the main text for comparison, both having very similar $\mathrm{var}(\Delta Q(V)_\mathrm{w3-w0})$ values. 

\begin{figure}[h!]
	\includegraphics[width=.8\textwidth]{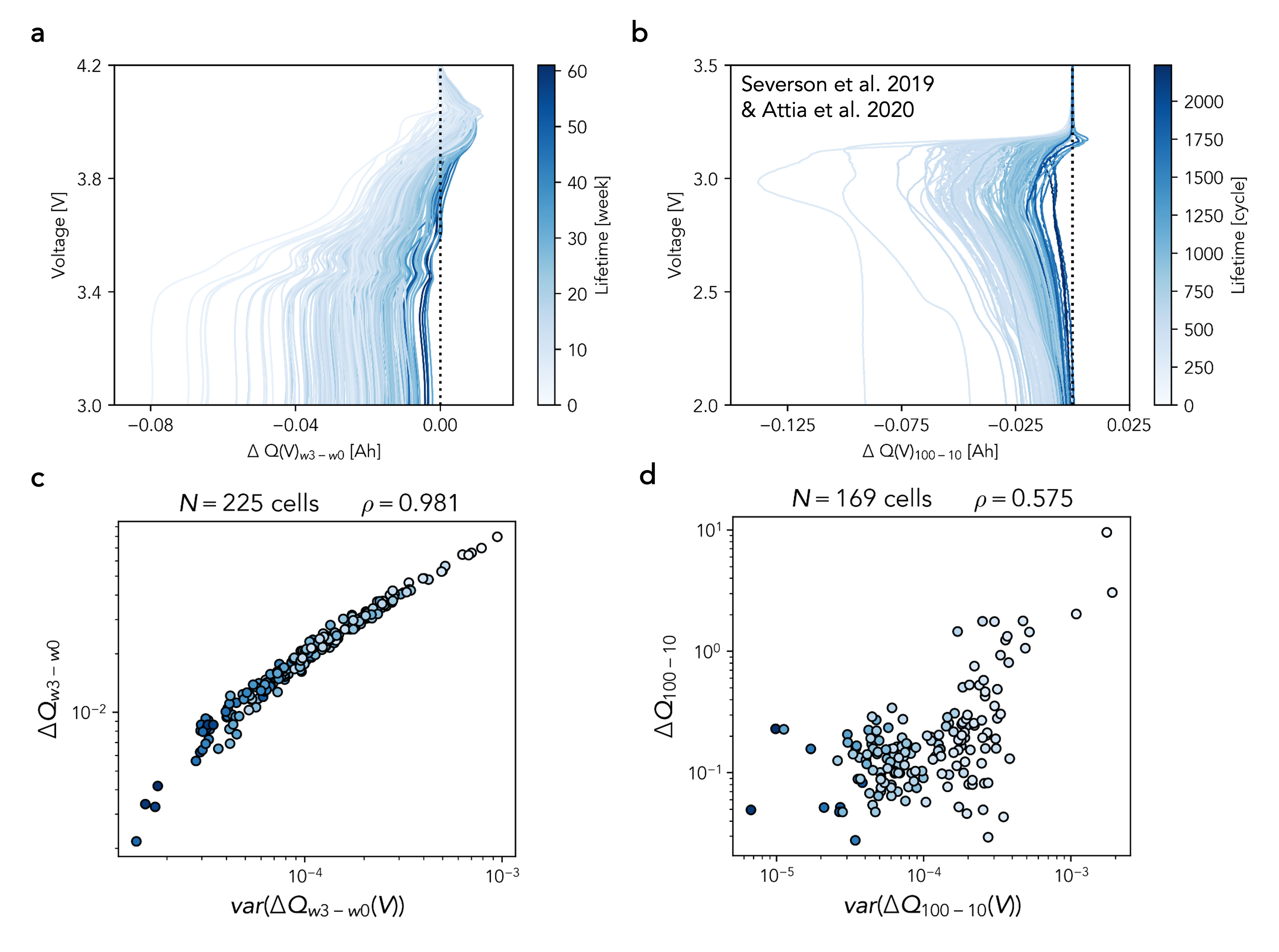}
	\centering
	\caption{An overview of $\Delta Q(V)$ curves from cells in two aging datasets. \textbf{a}, the ISU-ILCC battery aging dataset. \textbf{b}, the dataset of \citep{severson2019data,attia2020closed}.\textbf{c}, $\mathrm{var}\left(\Delta Q(V)_{\mathrm{w}3-\mathrm{w}0}\right)$ of the ISU-ILCC dataset. \textbf{d}, $\mathrm{var}\left(\Delta Q(V)_{100-10}\right)$ of the dataset of \citep{severson2019data,attia2020closed}. }
	\label{fig:deltaQ_curves_comparison}
	\centering
 \end{figure}

To further understand the reason behind the unsatisfactory performance of the well-known feature extracted from $\Delta Q(V)$ curves, we plot the $\Delta Q(V)_\mathrm{w3-wo}$ curves of the ISU-ILCC dataset (see Fig. \ref{fig:deltaQ_curves_comparison}a), along with the $\Delta Q(V)_\mathrm{100-10}$ from the dataset of \citep{severson2019data,attia2020closed} (see Fig. \ref{fig:deltaQ_curves_comparison}b). Apparently, the maximum difference exists at the lower cutoff voltage ($V=\SI{3.0}{V}$), which also reflects the total capacity fade between weeks 0 and 3. So, as shown in Fig. \ref{fig:deltaQ_curves_comparison}c, the $\mathrm{var}(\Delta Q(V)_\mathrm{w3-w0})$ correlates with the capacity fade. Both phenomena indicate that the adapted feature on our dataset mainly captures the capacity loss instead of the information related to early-stage degradation mechanisms. 

\subsection*{Analysis of Incremental Capacity Feature}
\begin{figure}
	\includegraphics[width=1\textwidth]{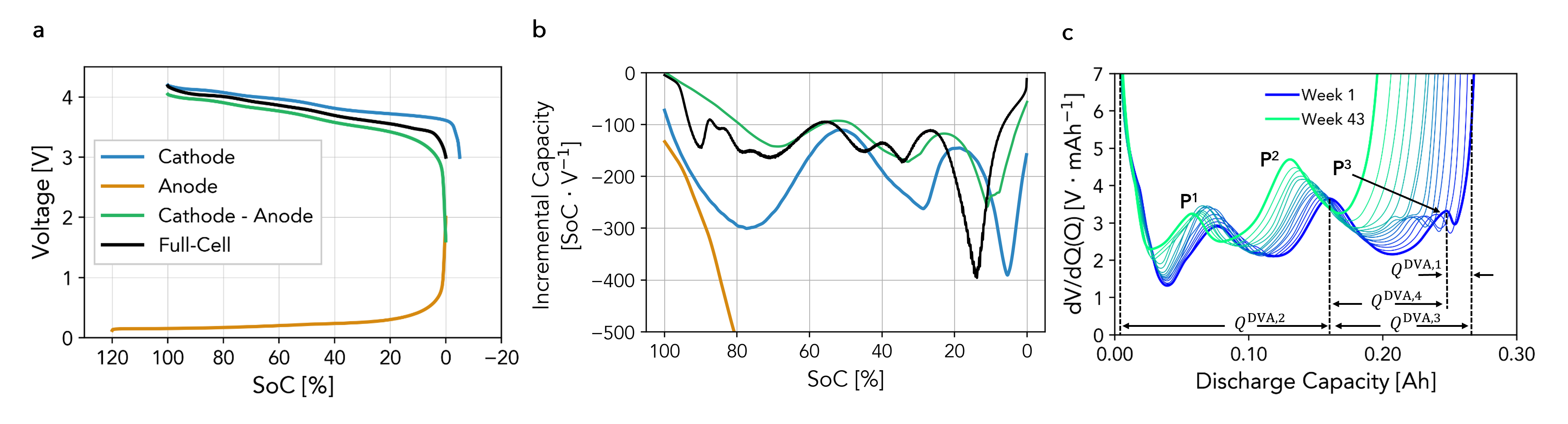}
	\centering
	\caption{An overview of the peak identification and tracking method for differential voltage feature engineering. \textbf{a}, Experimentally obtained positive and negative half-cell voltage as a function of the state of charge, their difference, and a full-cell curve for comparison. \textbf{b}, the incremental capacity as a function of the state of charge for each curve in \textbf{a}. The observable peaks in the half-cell curves indicate which electrode they originate from. \textbf{c}, Differential voltage as a function of cell capacity for cell one from group one (G1C1), illustrating the change in cell capacity and peak location during aging.}
	\label{fig:dva_extraction}
	\centering
\end{figure}

To clarify the relationship between the peaks in the differential voltage curve and cell health, we constructed half-cells from electrode materials obtained from disassembling a fresh cell. We cycled the half-cells at a slow rate (C/20) and reconstructed a full-cell pseudo-open circuit voltage curve. The results presented in Fig.\ \ref{fig:dva_extraction}a. However, the negative electrode data is poor because half-cell assembly was challenging. During assembly, we had to remove a water-soluble coating covering the negative electrode material by scratching it off, as using solvents would have damaged it. This process is inexact, and it produced poor electrode material, which then yielded poor results during cycling. Thus, we were unable to attribute peaks on incremental capacity curves to the specific side of electrodes. Future work on exploring degradation mechanisms in this dataset and the relationship between the early life features and the dominant degradation modes will help us determine which electrode this feature corresponds to.

The remaining unexplained variance in the new feature-lifetime correlation is likely due to the unavoidable influence of a decreasing lithium inventory on the shape of the $dQ/dV(V)$ curves. Decreases in lithium inventory can cause shifts in the voltages where peaks occur \citep{dubarry2022best}. This causes a small misalignment between the curves at weeks three and zero that varies cell-to-cell and introduces variation in the incremental capacity feature extraction. Destructively analyzing specific cells from the dataset would help to determine more concretely what the new feature $\mathrm{mean}\left(\Delta dQ/dV_{\mathrm{w3}-\mathrm{w0}}^{3.60\mathrm{V}-3.90\mathrm{V}}(V)\right)$ is capturing, but this was outside the current scope. 

\subsection*{Differential Voltage Features}
\label{sec:dva_feature_extraction}

Like incremental capacity, differential voltage $(dV/dQ(Q))$ analysis can effectively diagnose different component-level degradation modes in Li-ion cells \citep{raj2020investigation, dahn2012user, li2019degradation}. However, differential voltage analysis has yet to be widely used as part of automated feature extraction methods because curve manipulation and automatic peak detection are challenging. Unlike incremental capacity, the differential voltage is defined as a function of cell capacity, which can change cycle-to-cycle. The changing capacity makes curve manipulation and feature extraction via vector operations more difficult, as any two curves will not be the same length. Furthermore, the peaks and valleys of cells experiencing fast degradation often merge, confusing maxima and minima detection algorithms. 

Despite these challenges, we investigated extracting four capacity-based features from differential voltage curves. The four features, $Q^{\mathrm{DVA,1}}$ to $Q^{\mathrm{DVA,4}}$ in Fig.\ \ref{fig:dva_extraction}c, are designed to capture the evolution of the differential voltage curve during early life and are derived from the locations of peaks. The features capture the rate of change of different capacities and the relative shifts in the differential voltage curves, calculated as $\Delta Q_{\mathrm{w3}-\mathrm{w0}}^{\mathrm{DVA,1}} = Q_{\mathrm{w3}}^{\mathrm{DVA,1}} - Q_{\mathrm{w0}}^{\mathrm{DVA,1}}$. 

The four differential voltage features are designed to quantify capacity losses attributed to each electrode and capture shifts in the relative electrode balancing \citep{keil2016calendar}. Keil et al.\ \citep{keil2016calendar} suggest certain capacities can be estimated to determine the change in electrode balancing and the loss of active materials at the positive and negative electrodes ($\mathrm{LAM_{PE}}$ and $\mathrm{LAM_{NE}}$, respectively). A change in the cathode capacity is captured through $Q^{\mathrm{DVA,2}}$ since all the features of interest in this range are cathode specific. Similarly, the anode capacity is captured through $Q^{\mathrm{DVA,3}}$. The different balances of the two electrodes are captured through $Q^{\mathrm{DVA,1}}$ and $Q^{\mathrm{DVA,4}}$, tracking the anode and cathode peaks, respectively. Each of the four features is included in the feature selection process.

\subsection*{Constant Voltage Charging Times Feature}
\label{sec:CV_features}
In addition to features extracted from capacity-voltage curves and their derivatives, we derive a set of features from direct cell measurements of time and capacity. A benefit of these features is that they can be achieved using lower sampling frequency, measurement precision, and less data processing than the aforementioned curve difference features, making them suitable for implementation on battery health monitoring devices. The first feature extracted is the time spent in the constant-voltage (CV) charging step during each RPT, denoted $\mathrm{CV\; Time}_{wi}$. We also calculate the difference between two weeks' constant-voltage charging times, denoted $\Delta \mathrm{CV\; Time}_{\mathrm{w3}-\mathrm{w0}}$. A panel plot illustrating the extraction of these features and their correlation with cell lifetimes is included in the Supplementary Information. During charging, the constant-voltage charging step occurs as the final stage in charging. Data collected from CV charging steps have successfully been used to estimate the state of health of Li-ion batteries in recent literature \citep{EDDAHECH2014determine_soh_cv, YANG2018online_soh_cv, WANG2019soh_cv}. The extracted CV features reflect the interaction between capacity loss (decreasing the overall charging time) and increasing resistance to intercalation due to the degradation of the active electrode material. Additionally, we extracted features from the discharge capacity in RPTs, such as the cells' initial capacity $Q_{\mathrm{w0}}$ and the capacity fade between weeks three and zero $\Delta Q_{\mathrm{w3}-\mathrm{w0}}$, capturing the initial state of the cell and its relative change during early life, respectively.

\subsection*{Feature Extraction Details}
The following table contains all extracted features and their Pearson linear correlation coefficient with $\log(lifetime)$. Additionally, we mark the specific features selected for use in the degradation-informed models using the values 2 and 3, corresponding to the elastic net and HBM models built using two and three features, respectively.

\renewcommand{\arraystretch}{1.5}

\begin{table}[H]
\label{table:feature_list}
\centering
\caption{All extracted early-life features. Includes both condition-level and cell-level features.}
\begin{adjustbox}{max width=\textwidth}
\begin{tabular}{lrcc} 
\toprule
\textbf{Feature} &$\bm{\rho}$\textbf{ with log(lifetime)} &\textbf{Selected for Elastic Net} & \textbf{Selected for HBM}\\ 
\hline
$C_{\mathrm{chg}}$ & $-0.178$\\
$C_{\mathrm{dchg}}$ & $-0.086$\\
$\mathrm{DoD}$& $-0.682$ & 3 & Cell-level: 3\\
${C_{\mathrm{chg}}}^{0.5}DoD^{0.5}$&$-0.689$\\
${C_{\mathrm{dchg}}}^{0.5}DoD^{0.5}$&$-0.543$\\
${C_{\mathrm{chg}}}^{0.5}DoD^{0.5}+{C_{\mathrm{dchg}}}^{0.5}DoD^{0.5}$ & $-0.784$ & &Clustering Variable: 2, 3\\
$\log (\left|mean(\Delta dQ/dV_{\mathrm{w3}-\mathrm{w0}}(V)\right|)$&$-0.668$\\
$\log (\left|var(\Delta dQ/dV_{\mathrm{w3}-\mathrm{w0}}(V)\right|)$&$-0.634$\\
$\log \left(\left|mean(\Delta dQ/dV_{\mathrm{w3}-\mathrm{w0}}^{3.0V-3.6V}(V)\right|\right)$ & $-0.143$\\
$\log (\left|mean(\Delta dQ/dV_{\mathrm{w3}-\mathrm{w0}}^{3.6V-3.9V}(V)\right|)$& $-0.848$& 2, 3 & Cell-level: 2, 3 \\
$\log (\left|mean(\Delta dQ/dV_{\mathrm{w3}-\mathrm{w0}}^{3.9V-4.2V}(V)\right|)$&$-0.097$ \\
$\log (\left|var(\Delta dQ/dV_{\mathrm{w3}-\mathrm{w0}}^{3.0V-3.6V}(V)\right|)$&$-0.367$\\
$\log (\left|var(\Delta dQ/dV_{\mathrm{w3}-\mathrm{w0}}^{3.6V-3.9V}(V)\right|)$&$-0.315$\\
$\log (\left|var(\Delta dQ/dV_{\mathrm{w3}-\mathrm{w0}}^{3.9V-4.2V}(V)\right|)$&$-0.482$\\
$\log (\left|mean(\Delta Q_{\mathrm{w3}-\mathrm{w0}}(V)\right|)$&$-0.716$\\
$\log (\left|var(\Delta Q_{\mathrm{w3}-\mathrm{w0}}(V)\right|)$&$-0.686$\\
$\log (\mathrm{CV\;Time}_{\mathrm{w0}})$&$0.223$\\
$\log (\mathrm{CV\;Time}_{\mathrm{w3}})$&$-0.141$\\
$\log (\left|\Delta \mathrm{CV\;Time}_{\mathrm{w3}-\mathrm{w0}}\right|)$& $0.369$ & 2, 3 & Cell-level: 2, 3 \\
$\log(Q_{\mathrm{w0}})$&$-0.048$\\
$\log(Q_{\mathrm{w0}}^{3.0V-3.6V})$&$0.241$\\
$\log(Q_{\mathrm{w0}}^{3.6V-3.9V})$&$0.103$\\
$\log(Q_{\mathrm{w0}}^{3.9V-4.2V})$&$-0.278$\\
$\log(\Delta Q_{\mathrm{w3}-\mathrm{w0}})$ & $-0.668$\\
$\Delta Q_{\mathrm{w3}-\mathrm{w0}}^{1}$&$0.096$\\
$\Delta Q_{\mathrm{w3}-\mathrm{w0}}^{2}$&$-0.567$\\
$\Delta Q_{\mathrm{w3}-\mathrm{w0}}^{3}$&$-0.473$\\
$\Delta Q_{\mathrm{w3}-\mathrm{w0}}^{4}$&$-0.409$\\

\bottomrule
\end{tabular}
\end{adjustbox}

\end{table}

\renewcommand{\arraystretch}{1}

\clearpage
\subsection*{Selected Feature Details}

\begin{figure}[h!]
	\includegraphics[width=0.88\textwidth]{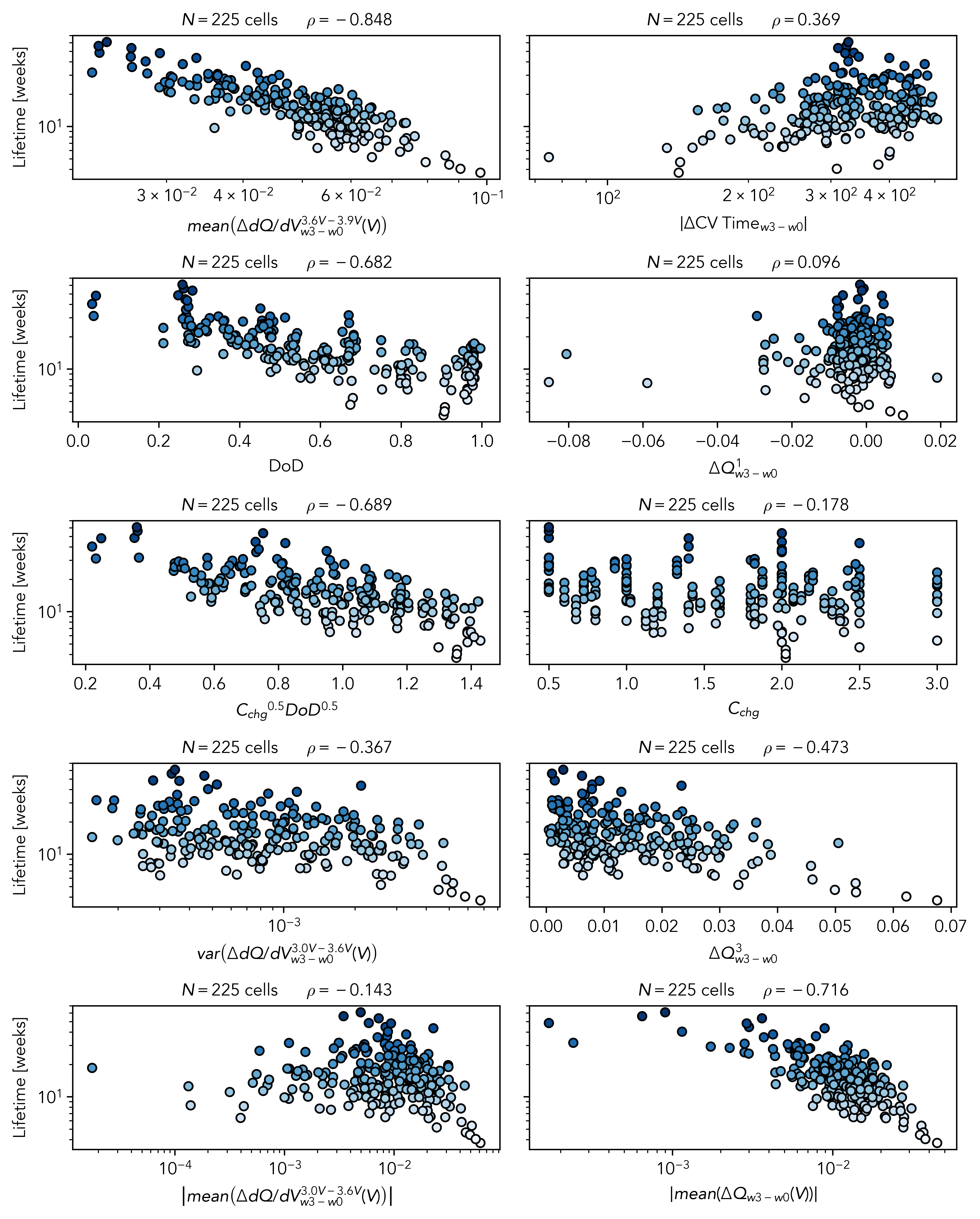}
	\centering
	\caption{An overview of the 10 selected features based on the results from the step-wise feature selection technique.}
	\label{fig:selected_10_features}
	\centering
 \end{figure}

\clearpage

\subsection*{Time Dependence of Input Features}

\begin{figure}
	\includegraphics[width=1.0\textwidth]{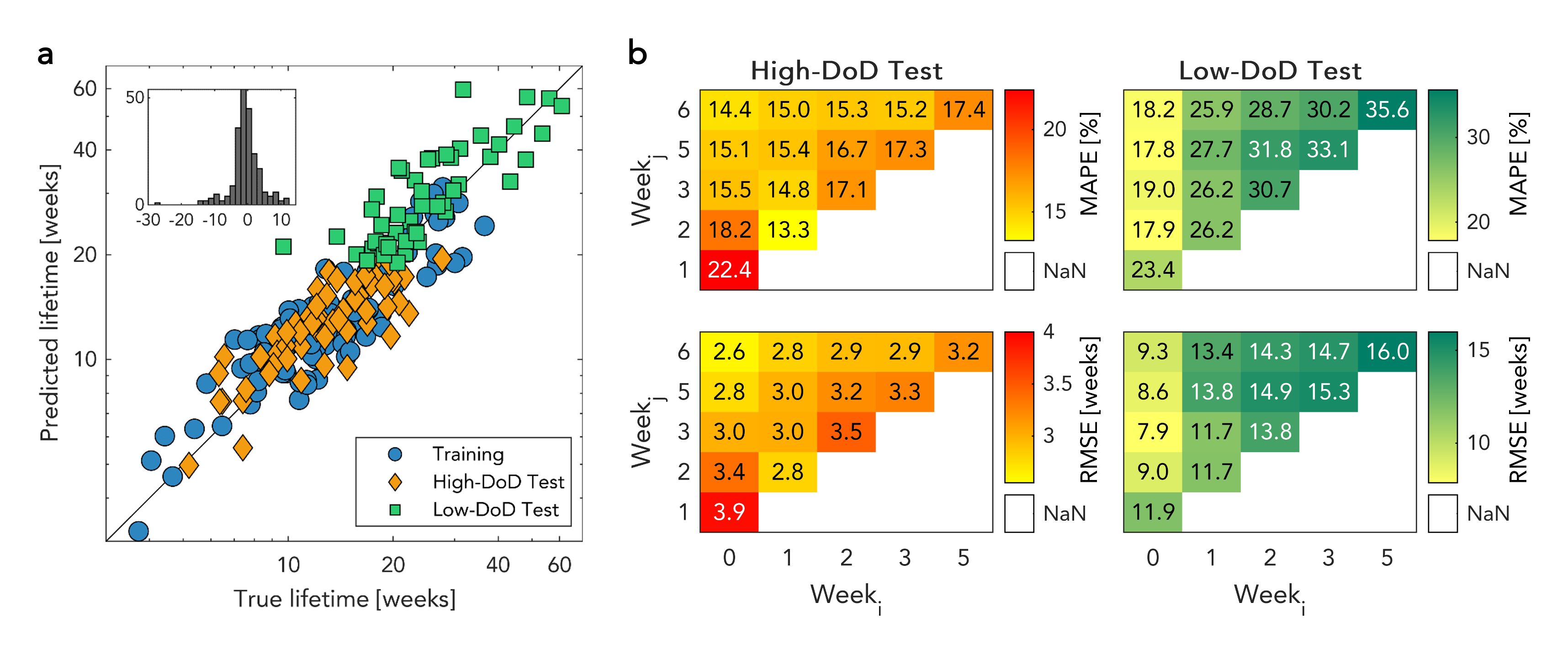}
	\centering
	\caption{Overview of prediction results for \emph{degradation-informed} elastic net model with two input features. \textbf{a}, True and predicted lifetimes using features extracted from weeks three and zero $(\mathrm{w_3}-\mathrm{w_0})$ with embedded histogram showing prediction residuals. \textbf{b}, The $\mathrm{RMSE_{EOL}}$ and $\mathrm{MAPE_{EOL}}$ error metrics as a function of week numbers from which the early-life features, denoted $(w_j-w_i)$. Week four data is omitted from the study due to a testing error.}
	\label{fig:truepredicted}
	\centering
 \end{figure}

Motivated by the need to predict cell lifetime as early as possible, we performed a study varying the time frame from which the features are extracted to understand the impact on model accuracy. Unfortunately, a testing error during week four caused irreversible data loss for a large batch of cells, so week four data is omitted from this study. Using the degradation-informed elastic net model with two features, we vary the RPTs from which the features are extracted and record the test errors for the high and low-DoD test datasets. The results are shown in Fig.\ \ref{fig:truepredicted}b.

First, we analyze the accuracy trend with the starting week fixed to week zero (i.e., $w_i=0$). Under this setting, the prediction errors on the high-DoD dataset consistently decrease as the time between RPTs increases. However, the prediction errors for the low-DoD test set are found to slightly increase with increasing time-frame around weeks five and six $w_j = 5,6$. This is likely because many cells experience rapid degradation after week five/six, which alters the feature-lifetime relationship for cells with short lives. This causes the model to change its fit, decreasing its prediction accuracy on long-lifetime cells. 

Second, we analyze the accuracy trend by looking at the time between any two RPTs. Along the diagonal, the delta between any two RPTs is one week. Under these conditions, we observe a substantial increase in model prediction error on both the high- and low-DoD test sets compared with counterparts toward the upper left corner (i.e., models with features extracted with intervals longer than one week). This suggests a minimum time interval of $(w_j -w_i) \ge 2$ is required to accurately estimate the rate of degradation inside the cell from early-life features.

Finally, we observe that the model prediction error on the low-DoD test set continuously increases with increasing starting week $w_i$. This could be an effect of optimizing the incremental capacity feature using data from weeks three and zero. The optimal voltage range for this feature may change with the RPTs used and was not accounted for in this study.

\subsection*{Training process of hierarchical Bayesian model}
The overall training process follows:

\begin{itemize}
\item Estimate level-2 posterior distribution by Bayes's rule 

\begin{equation*}
    P\left(\bm{\gamma}, \sigma \mid\left\{Y\right\}\right) \propto \prod_{j=1}^{J} P\left(Y_{j}\mid \bm{\gamma},\sigma \right) \cdot P(\bm{\gamma},\sigma)
\end{equation*}

\item Use level-2 posterior distribution as prior for level-1 parameters, calculate level-1 posterior distribution 
\begin{equation*}
    P\left(\bm{\theta_j}, \sigma_{j} \mid \bm{\gamma},\sigma, Y_j\right)=\frac{P\left(Y_j \mid \bm{\theta_j},\sigma_j\right) \cdot P\left(\bm{\theta_j},\sigma_j \mid \bm{\gamma},\sigma\right) \cdot P(\bm{\gamma},\sigma \mid \left\{Y\right\})}{P\left(Y_j \mid \bm{\gamma},\sigma\right)}
\end{equation*}

\item Use level-1 posterior distribution to make predictions on individual labels
\begin{equation*}
    P\left(y_j^*\right)=\int_{\theta_j,\sigma_j} P\left(y_j^* \mid \bm{\theta_j}, \sigma_j\right) P\left(\bm{\theta_j},\sigma_j \mid \bm{\gamma}, Y_j\right) d \bm{\theta_j} d \sigma_j
\end{equation*}
\end{itemize}

\subsection*{HBM related analysis}

\begin{figure}
	\includegraphics[width=0.75\textwidth]{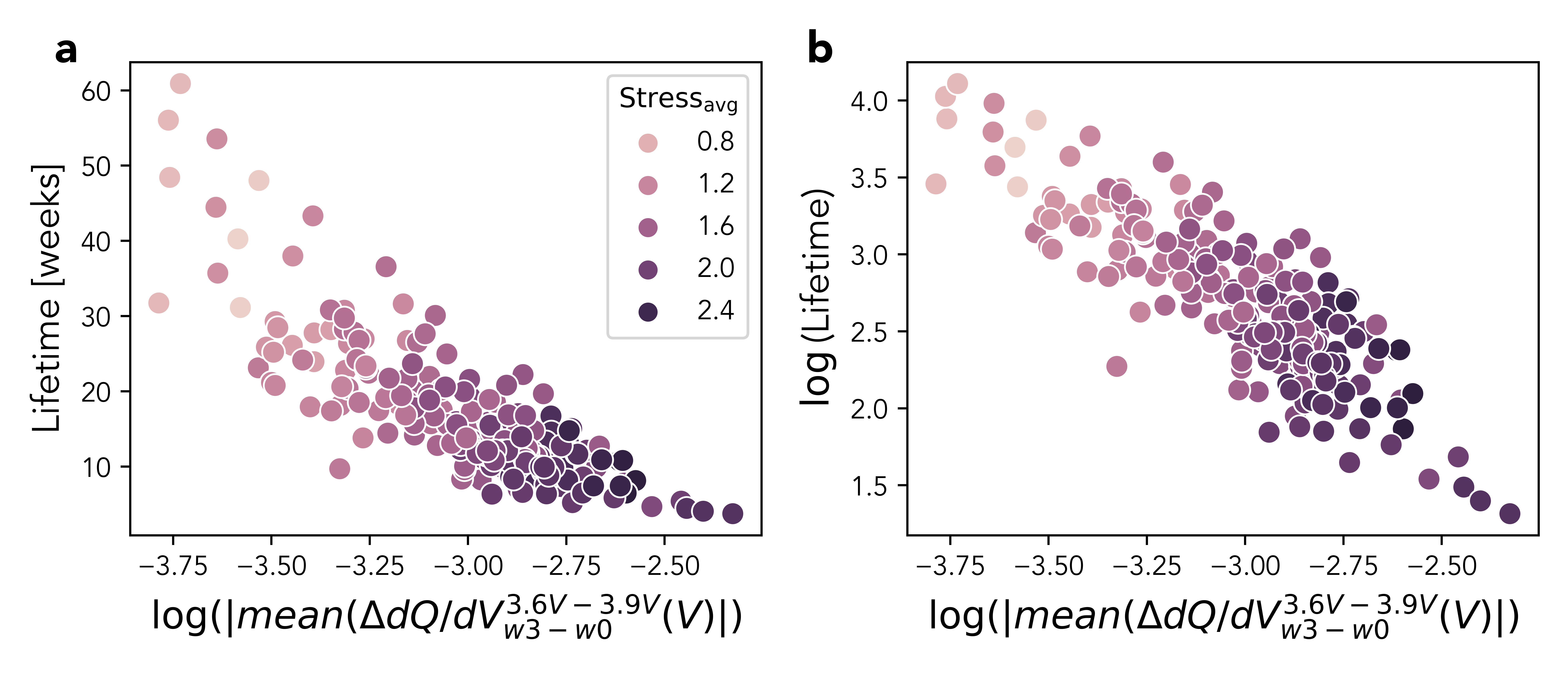}
	\centering
	\caption{Relationship between cell-level feature $\text{mean}\left(\Delta dQ/dV_{\mathrm{w3}-\mathrm{w0}}^{3.60\mathrm{V}-3.90\mathrm{V}}(V)\right)$ and cell lifetime, colored by $\mathrm{Stress_{\mathrm{avg}}}$. 
 }
	\label{fig:temp_con_ind}
	\centering
\end{figure}

 Fig.\ \ref{fig:temp_con_ind} shows a scatter plot of the cell-level feature $\text{mean}\left(\Delta dQ/dV_{\mathrm{w3}-\mathrm{w0}}^{3.60\mathrm{V}-3.90\mathrm{V}}(V)\right)$ vs.\ lifetime, colored by $\mathrm{Stress_{\mathrm{avg}}}$. 
 
As $\mathrm{Stress_{\mathrm{avg}}}$ decreases, the slope of the cell-level feature-lifetime relationship becomes steeper. However, the changing trend is not as clear when analyzing the data on a log-log plot. Normally, the reason for using the log transform on both the feature and the target is to increase the Pearson linear correlation coefficient, as a higher linear correlation will generally improve model prediction performance. This pursuit of a one-for-all linear relationship between the feature and the target hides the data's differences and hierarchical structure caused by the various cycling conditions. 

\begin{figure}
	\includegraphics[width=0.9\textwidth]{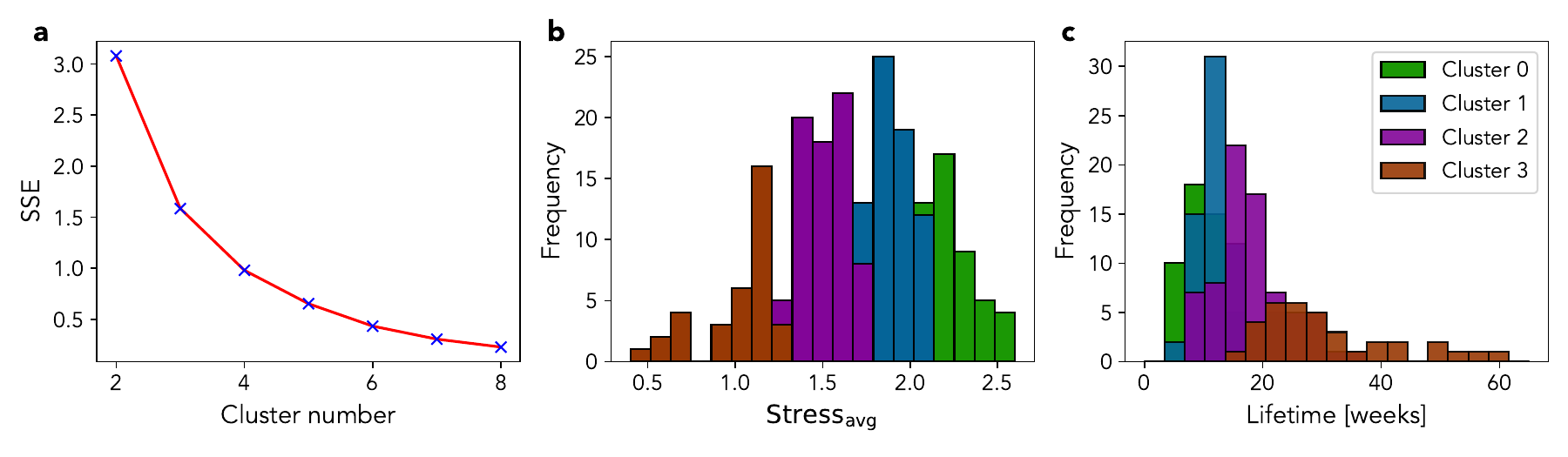}
	\centering
	\caption{Overview of clustering results. \textbf{a},  Influence of number of clusters on clustering score $\mathrm{SSE}$. \textbf{b}, Histogram of  stress factor $\mathrm{Stress_{\mathrm{avg}}}$ colored by cluster. \textbf{c}  Corresponding lifetime distribution for each cluster.}
	\label{fig:temp_clustering}
	\centering
\end{figure}

The clustering score $\mathrm{SSE} = \sum_{i=0}^N\left(x_i-c_i\right)^2$, which describes the sum of squared distances between sample points and their assigned centroid, is used to evaluate the influence of the number of clusters on the clustering results.

Fig.\ \ref{fig:temp_clustering}a shows the $\mathrm{SSE}$ as a function of the number of clusters. According to the empirical elbow rule \citep{yuan2019research}, we select $K=4$ clusters. From Figs.\ \ref{fig:temp_clustering}b and \ref{fig:temp_clustering}c, we observe two sources of variability that affect lifetimes. The first is the cross-cluster lifetime variability, which arises from differences in usage, and is measured as a difference in $\mathrm{Stress_{\mathrm{avg}}}$. The other source of lifetime variability arises from in-cluster differences due to manufacturing variability and cycling tester variability.

\begin{figure}
	\includegraphics[width=1\textwidth]{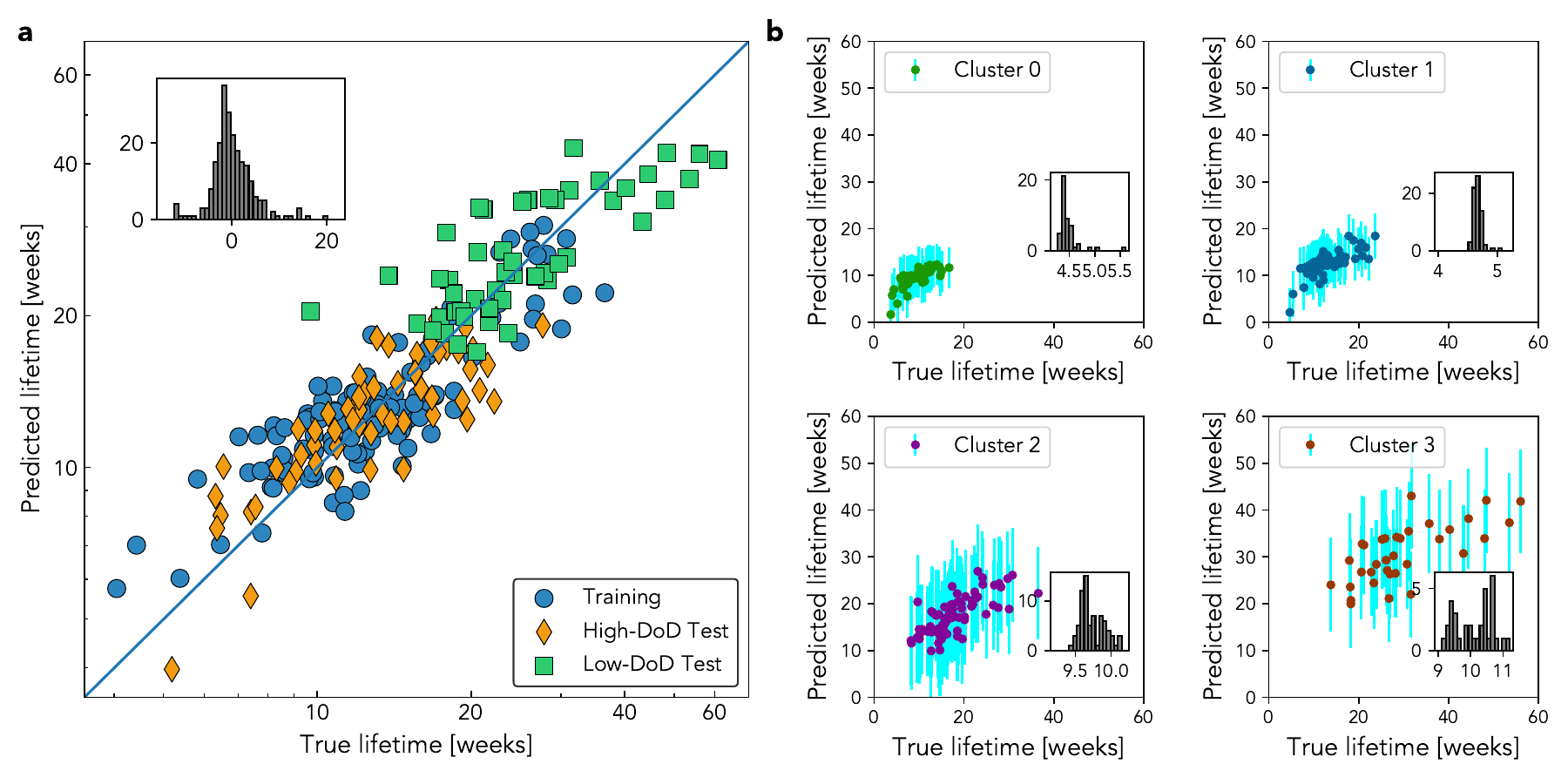}
	\centering
	\caption{Overview of HBM results. \textbf{a}, True vs.\ predicted lifetimes using the optimal two features extracted from weeks three and zero $(\mathrm{w_3}-\mathrm{w_0})$, with embedded histogram showing prediction residuals. \textbf{b}, Predictions for each cluster with 2 standard deviations as the corresponding error bar for each sample. The embedded histograms show a summary of error bars}
	\label{fig:hbm_results}
 \end{figure}

 \begin{figure}
	\includegraphics[width=1\textwidth]{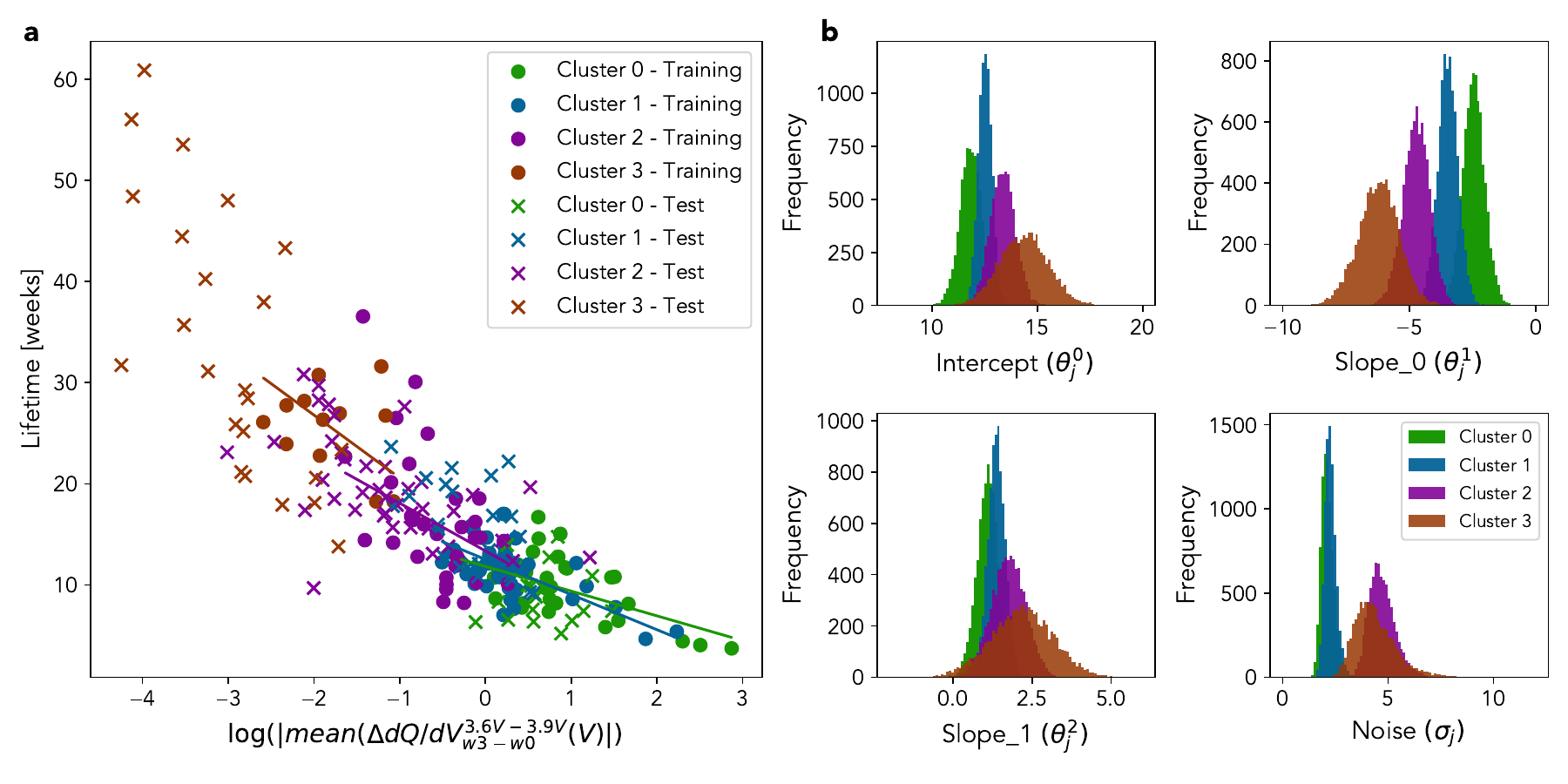}
	\centering
	\caption{Visualization of HBM regression parameter uncertainties. \textbf{a},  Relationship between $\log (\left|\text{mean}(\Delta dQ/dV_{\mathrm{w3}-\mathrm{w0}}^{3.6V-3.9V}(V)\right|)$ and true lifetime across different clusters and train-test split ("Test" denotes samples from both high- and low-DoD sets). Fits, corresponding to mean parameter values, are plotted for each cluster. \textbf{b}, Uncertainty histograms for regression parameters ($\theta_{j}^0, \theta_{j}^1, \theta_{j}^2, \sigma_{j}$) of each cluster.}
	\label{fig:hbm_detail_uncertainty}
	\centering
 \end{figure}

Further analysis of uncertainty is shown in Fig.\ \ref{fig:hbm_detail_uncertainty}. The HBM successfully captures the changing slope describing the relationship between  $\log (\left|\text{mean}(\Delta dQ/dV_{\mathrm{w3}-\mathrm{w0}}^{3.6V-3.9V}(V)\right|)$ and true lifetime in Fig.\ \ref{fig:hbm_detail_uncertainty}a. By exploiting the assumption that cell-level regression coefficients are decided by cycling stress cluster-level features, the HBM gives a reasonable fit for Cluster 3 samples (46) based on a very limited training set (12). Considering the posterior predictive distribution expression $p\left(y_j^* \mid Y_j\right)$, the uncertainty on predictions is influenced by both the uncertainly from the regression intercepts and slopes $\theta_j$, and the uncertainty due to measurement noises $\sigma_j$. Fig.\ \ref{fig:hbm_detail_uncertainty}b shows these two kinds of uncertainties across all clusters. The posterior probability distributions for $\bm{\theta_j}$ and $\sigma_j$ are much wider for cluster 3 than for any other clusters.

\bibliographystyle{unsrtnat}
\bibliography{REFERENCES}  

\begin{thebibliography}{56}
\providecommand{\natexlab}[1]{#1}
\providecommand{\url}[1]{\texttt{#1}}
\expandafter\ifx\csname urlstyle\endcsname\relax
  \providecommand{\doi}[1]{doi: #1}\else
  \providecommand{\doi}{doi: \begingroup \urlstyle{rm}\Url}\fi

\bibitem[Birkl et~al.(2017)Birkl, Roberts, McTurk, Bruce, and Howey]{birkl2017degradation}
Christoph~R Birkl, Matthew~R Roberts, Euan McTurk, Peter~G Bruce, and David~A Howey.
\newblock Degradation diagnostics for lithium ion cells.
\newblock \emph{Journal of Power Sources}, 341:\penalty0 373--386, 2017.

\bibitem[Sulzer et~al.(2021)Sulzer, Mohtat, Aitio, Lee, Yeh, Steinbacher, Khan, Lee, Siegel, Stefanopoulou, et~al.]{sulzer2021challenge}
Valentin Sulzer, Peyman Mohtat, Antti Aitio, Suhak Lee, Yen~T Yeh, Frank Steinbacher, Muhammad~Umer Khan, Jang~Woo Lee, Jason~B Siegel, Anna~G Stefanopoulou, et~al.
\newblock The challenge and opportunity of battery lifetime prediction from field data.
\newblock \emph{Joule}, 5\penalty0 (8):\penalty0 1934--1955, 2021.

\bibitem[Thelen et~al.(2022)Thelen, Lui, Shen, Laflamme, Hu, Ye, and Hu]{thelen2022integrating}
Adam Thelen, Yu~Hui Lui, Sheng Shen, Simon Laflamme, Shan Hu, Hui Ye, and Chao Hu.
\newblock Integrating physics-based modeling and machine learning for degradation diagnostics of lithium-ion batteries.
\newblock \emph{Energy Storage Materials}, 2022.

\bibitem[Kunz et~al.(2021)Kunz, Dufek, Yi, Gering, Shirk, Smith, Chen, Wang, Gasper, Bewley, et~al.]{kunz2021early}
M~Ross Kunz, Eric~J Dufek, Zonggen Yi, Kevin~L Gering, Matthew~G Shirk, Kandler Smith, Bor-Rong Chen, Qiang Wang, Paul Gasper, Randy~L Bewley, et~al.
\newblock Early battery performance prediction for mixed use charging profiles using hierarchal machine learning.
\newblock \emph{Batteries \& Supercaps}, 4\penalty0 (7):\penalty0 1186--1196, 2021.

\bibitem[Severson et~al.(2019)Severson, Attia, Jin, Perkins, Jiang, Yang, Chen, Aykol, Herring, Fraggedakis, et~al.]{severson2019data}
Kristen~A Severson, Peter~M Attia, Norman Jin, Nicholas Perkins, Benben Jiang, Zi~Yang, Michael~H Chen, Muratahan Aykol, Patrick~K Herring, Dimitrios Fraggedakis, et~al.
\newblock Data-driven prediction of battery cycle life before capacity degradation.
\newblock \emph{Nature Energy}, 4\penalty0 (5):\penalty0 383--391, 2019.

\bibitem[Attia et~al.(2020)Attia, Grover, Jin, Severson, Markov, Liao, Chen, Cheong, Perkins, Yang, et~al.]{attia2020closed}
Peter~M Attia, Aditya Grover, Norman Jin, Kristen~A Severson, Todor~M Markov, Yang-Hung Liao, Michael~H Chen, Bryan Cheong, Nicholas Perkins, Zi~Yang, et~al.
\newblock Closed-loop optimization of fast-charging protocols for batteries with machine learning.
\newblock \emph{Nature}, 578\penalty0 (7795):\penalty0 397--402, 2020.

\bibitem[Dave et~al.(2020)Dave, Mitchell, Kandasamy, Wang, Burke, Paria, P{\'o}czos, Whitacre, and Viswanathan]{dave2020autonomous}
Adarsh Dave, Jared Mitchell, Kirthevasan Kandasamy, Han Wang, Sven Burke, Biswajit Paria, Barnab{\'a}s P{\'o}czos, Jay Whitacre, and Venkatasubramanian Viswanathan.
\newblock Autonomous discovery of battery electrolytes with robotic experimentation and machine learning.
\newblock \emph{Cell Reports Physical Science}, 1\penalty0 (12):\penalty0 100264, 2020.

\bibitem[Smith et~al.(2009)Smith, Burns, Trussler, and Dahn]{smith2009precision}
AJ~Smith, JC~Burns, S~Trussler, and JR~Dahn.
\newblock Precision measurements of the coulombic efficiency of lithium-ion batteries and of electrode materials for lithium-ion batteries.
\newblock \emph{Journal of The Electrochemical Society}, 157\penalty0 (2):\penalty0 A196, 2009.

\bibitem[Burns et~al.(2011)Burns, Jain, Smith, Eberman, Scott, Gardner, and Dahn]{burns2011evaluation}
JC~Burns, Gaurav Jain, AJ~Smith, KW~Eberman, Erik Scott, JP~Gardner, and JR~Dahn.
\newblock Evaluation of effects of additives in wound li-ion cells through high precision coulometry.
\newblock \emph{Journal of The Electrochemical Society}, 158\penalty0 (3):\penalty0 A255, 2011.

\bibitem[Burns et~al.(2013)Burns, Kassam, Sinha, Downie, Solnickova, Way, and Dahn]{burns2013predicting}
JC~Burns, Adil Kassam, NN~Sinha, LE~Downie, Lucie Solnickova, BM~Way, and JR~Dahn.
\newblock Predicting and extending the lifetime of li-ion batteries.
\newblock \emph{Journal of The Electrochemical Society}, 160\penalty0 (9):\penalty0 A1451, 2013.

\bibitem[Baumh{\"o}fer et~al.(2014)Baumh{\"o}fer, Br{\"u}hl, Rothgang, and Sauer]{baumhofer2014production}
Thorsten Baumh{\"o}fer, Manuel Br{\"u}hl, Susanne Rothgang, and Dirk~Uwe Sauer.
\newblock Production caused variation in capacity aging trend and correlation to initial cell performance.
\newblock \emph{Journal of Power Sources}, 247:\penalty0 332--338, 2014.

\bibitem[Harris et~al.(2017)Harris, Harris, and Li]{harris2017failure}
Stephen~J Harris, David~J Harris, and Chen Li.
\newblock Failure statistics for commercial lithium ion batteries: A study of 24 pouch cells.
\newblock \emph{Journal of Power Sources}, 342:\penalty0 589--597, 2017.

\bibitem[Greenbank and Howey(2021)]{greenbank2021automated}
Samuel Greenbank and David Howey.
\newblock Automated feature extraction and selection for data-driven models of rapid battery capacity fade and end of life.
\newblock \emph{IEEE Transactions on Industrial Informatics}, 18\penalty0 (5):\penalty0 2965--2973, 2021.

\bibitem[Zhang et~al.(2021)Zhang, Peng, Guan, and Wu]{zhang2021prognostics}
Yu~Zhang, Zhen Peng, Yong Guan, and Lifeng Wu.
\newblock Prognostics of battery cycle life in the early-cycle stage based on hybrid model.
\newblock \emph{Energy}, 221:\penalty0 119901, 2021.

\bibitem[Yang et~al.(2020)Yang, Wang, Xu, Huang, and Tsui]{yang2020lifespan}
Fangfang Yang, Dong Wang, Fan Xu, Zhelin Huang, and Kwok-Leung Tsui.
\newblock Lifespan prediction of lithium-ion batteries based on various extracted features and gradient boosting regression tree model.
\newblock \emph{Journal of Power Sources}, 476:\penalty0 228654, 2020.

\bibitem[Weng et~al.(2021)Weng, Mohtat, Attia, Sulzer, Lee, Less, and Stefanopoulou]{weng2021predicting}
Andrew Weng, Peyman Mohtat, Peter~M Attia, Valentin Sulzer, Suhak Lee, Greg Less, and Anna Stefanopoulou.
\newblock Predicting the impact of formation protocols on battery lifetime immediately after manufacturing.
\newblock \emph{Joule}, 5\penalty0 (11):\penalty0 2971--2992, 2021.

\bibitem[Saxena et~al.(2022)Saxena, Ward, Kubal, Lu, Babinec, and Paulson]{saxena2022convolutional}
Saurabh Saxena, Logan Ward, Joseph Kubal, Wenquan Lu, Susan Babinec, and Noah Paulson.
\newblock A convolutional neural network model for battery capacity fade curve prediction using early life data.
\newblock \emph{Journal of Power Sources}, 542:\penalty0 231736, 2022.

\bibitem[Herring et~al.(2020)Herring, Gopal, Aykol, Montoya, Anapolsky, Attia, Gent, Hummelsh{\o}j, Hung, Kwon, et~al.]{herring2020beep}
Patrick Herring, Chirranjeevi~Balaji Gopal, Muratahan Aykol, Joseph~H Montoya, Abraham Anapolsky, Peter~M Attia, William Gent, Jens~S Hummelsh{\o}j, Linda Hung, Ha-Kyung Kwon, et~al.
\newblock Beep: A python library for battery evaluation and early prediction.
\newblock \emph{SoftwareX}, 11:\penalty0 100506, 2020.

\bibitem[Fei et~al.(2021)Fei, Yang, Tsui, Li, and Zhang]{fei2021early}
Zicheng Fei, Fangfang Yang, Kwok-Leung Tsui, Lishuai Li, and Zijun Zhang.
\newblock Early prediction of battery lifetime via a machine learning based framework.
\newblock \emph{Energy}, 225:\penalty0 120205, 2021.

\bibitem[Ferm{\'\i}n-Cueto et~al.(2020)Ferm{\'\i}n-Cueto, McTurk, Allerhand, Medina-Lopez, Anjos, Sylvester, and Dos~Reis]{fermin2020identification}
Paula Ferm{\'\i}n-Cueto, Euan McTurk, Michael Allerhand, Encarni Medina-Lopez, Miguel~F Anjos, Joel Sylvester, and Goncalo Dos~Reis.
\newblock Identification and machine learning prediction of knee-point and knee-onset in capacity degradation curves of lithium-ion cells.
\newblock \emph{Energy and AI}, 1:\penalty0 100006, 2020.

\bibitem[Li et~al.(2021)Li, Sengupta, Dechent, Howey, Annaswamy, and Sauer]{li2021one}
Weihan Li, Neil Sengupta, Philipp Dechent, David Howey, Anuradha Annaswamy, and Dirk~Uwe Sauer.
\newblock One-shot battery degradation trajectory prediction with deep learning.
\newblock \emph{Journal of Power Sources}, 506:\penalty0 230024, 2021.

\bibitem[Paulson et~al.(2022)Paulson, Kubal, Ward, Saxena, Lu, and Babinec]{paulson2022feature}
Noah~H Paulson, Joseph Kubal, Logan Ward, Saurabh Saxena, Wenquan Lu, and Susan~J Babinec.
\newblock Feature engineering for machine learning enabled early prediction of battery lifetime.
\newblock \emph{Journal of Power Sources}, 527:\penalty0 231127, 2022.

\bibitem[Attia et~al.(2022)Attia, Bills, Planella, Dechent, Dos~Reis, Dubarry, Gasper, Gilchrist, Greenbank, Howey, et~al.]{attia2022knees}
Peter~M Attia, Alexander Bills, Ferran~Brosa Planella, Philipp Dechent, Goncalo Dos~Reis, Matthieu Dubarry, Paul Gasper, Richard Gilchrist, Samuel Greenbank, David Howey, et~al.
\newblock “knees” in lithium-ion battery aging trajectories.
\newblock \emph{Journal of The Electrochemical Society}, 169\penalty0 (6):\penalty0 060517, 2022.

\bibitem[Waldmann et~al.(2018)Waldmann, Hogg, and Wohlfahrt-Mehrens]{waldmann2018li}
Thomas Waldmann, Bj{\"o}rn-Ingo Hogg, and Margret Wohlfahrt-Mehrens.
\newblock Li plating as unwanted side reaction in commercial li-ion cells--a review.
\newblock \emph{Journal of Power Sources}, 384:\penalty0 107--124, 2018.

\bibitem[Han et~al.(2019)Han, Lu, Zheng, Feng, Li, Li, and Ouyang]{han2019review}
Xuebing Han, Languang Lu, Yuejiu Zheng, Xuning Feng, Zhe Li, Jianqiu Li, and Minggao Ouyang.
\newblock A review on the key issues of the lithium ion battery degradation among the whole life cycle.
\newblock \emph{ETransportation}, 1:\penalty0 100005, 2019.

\bibitem[Dos~Reis et~al.(2021)Dos~Reis, Strange, Yadav, and Li]{dos2021lithium}
Goncalo Dos~Reis, Calum Strange, Mohit Yadav, and Shawn Li.
\newblock Lithium-ion battery data and where to find it.
\newblock \emph{Energy and AI}, 5:\penalty0 100081, 2021.

\bibitem[Bole et~al.(2014)Bole, Kulkarni, and Daigle]{bole2014adaptation}
Brian Bole, Chetan~S Kulkarni, and Matthew Daigle.
\newblock Adaptation of an electrochemistry-based li-ion battery model to account for deterioration observed under randomized use.
\newblock In \emph{Annual Conference of the PHM Society}, 2014.

\bibitem[Saha et~al.(2008)Saha, Goebel, Poll, and Christophersen]{saha2008prognostics}
Bhaskar Saha, Kai Goebel, Scott Poll, and Jon Christophersen.
\newblock Prognostics methods for battery health monitoring using a bayesian framework.
\newblock \emph{IEEE Transactions on instrumentation and measurement}, 58\penalty0 (2):\penalty0 291--296, 2008.

\bibitem[He et~al.(2011)He, Williard, Osterman, and Pecht]{he2011prognostics}
Wei He, Nicholas Williard, Michael Osterman, and Michael Pecht.
\newblock Prognostics of lithium-ion batteries based on dempster--shafer theory and the bayesian monte carlo method.
\newblock \emph{Journal of Power Sources}, 196\penalty0 (23):\penalty0 10314--10321, 2011.

\bibitem[Xing et~al.(2013)Xing, Ma, Tsui, and Pecht]{xing2013ensemble}
Yinjiao Xing, Eden~WM Ma, Kwok-Leung Tsui, and Michael Pecht.
\newblock An ensemble model for predicting the remaining useful performance of lithium-ion batteries.
\newblock \emph{Microelectronics Reliability}, 53\penalty0 (6):\penalty0 811--820, 2013.

\bibitem[Preger et~al.(2020)Preger, Barkholtz, Fresquez, Campbell, Juba, Rom{\`a}n-Kustas, Ferreira, and Chalamala]{preger2020degradation}
Yuliya Preger, Heather~M Barkholtz, Armando Fresquez, Daniel~L Campbell, Benjamin~W Juba, Jessica Rom{\`a}n-Kustas, Summer~R Ferreira, and Babu Chalamala.
\newblock Degradation of commercial lithium-ion cells as a function of chemistry and cycling conditions.
\newblock \emph{Journal of The Electrochemical Society}, 167\penalty0 (12):\penalty0 120532, 2020.

\bibitem[Dechent et~al.(2021)Dechent, Greenbank, Hildenbrand, Jbabdi, Sauer, and Howey]{dechent2021estimation}
Philipp Dechent, Samuel Greenbank, Felix Hildenbrand, Saad Jbabdi, Dirk~Uwe Sauer, and David~A Howey.
\newblock Estimation of li-ion degradation test sample sizes required to understand cell-to-cell variability.
\newblock \emph{Batteries \& Supercaps}, 4\penalty0 (12):\penalty0 1821--1829, 2021.

\bibitem[Pastor-Fern{\'a}ndez et~al.(2017)Pastor-Fern{\'a}ndez, Uddin, Chouchelamane, Widanage, and Marco]{pastor2017comparison}
Carlos Pastor-Fern{\'a}ndez, Kotub Uddin, Gael~H Chouchelamane, W~Dhammika Widanage, and James Marco.
\newblock A comparison between electrochemical impedance spectroscopy and incremental capacity-differential voltage as li-ion diagnostic techniques to identify and quantify the effects of degradation modes within battery management systems.
\newblock \emph{Journal of Power Sources}, 360:\penalty0 301--318, 2017.

\bibitem[Berecibar et~al.(2016{\natexlab{a}})Berecibar, Dubarry, Omar, Villarreal, and Van~Mierlo]{berecibar2016degradation1}
Maitane Berecibar, Matthieu Dubarry, Noshin Omar, Igor Villarreal, and Joeri Van~Mierlo.
\newblock Degradation mechanism detection for nmc batteries based on incremental capacity curves.
\newblock \emph{World Electric Vehicle Journal}, 8\penalty0 (2):\penalty0 350--361, 2016{\natexlab{a}}.

\bibitem[Berecibar et~al.(2016{\natexlab{b}})Berecibar, Dubarry, Villarreal, Omar, and Van~Mierlo]{berecibar2016degradation2}
Maitane Berecibar, Matthieu Dubarry, Igor Villarreal, Noshin Omar, and Joeri Van~Mierlo.
\newblock Degradation mechanisms detection for hp and he nmc cells based on incremental capacity curves.
\newblock In \emph{2016 IEEE Vehicle Power and Propulsion Conference (VPPC)}, pages 1--5. IEEE, 2016{\natexlab{b}}.

\bibitem[Smith et~al.(2021)Smith, Gasper, Colclasure, Shimonishi, and Yoshida]{smith2021lithium}
Kandler Smith, Paul Gasper, Andrew~M Colclasure, Yuta Shimonishi, and Shuhei Yoshida.
\newblock Lithium-ion battery life model with electrode cracking and early-life break-in processes.
\newblock \emph{Journal of The Electrochemical Society}, 168\penalty0 (10):\penalty0 100530, 2021.

\bibitem[Reniers et~al.(2019)Reniers, Mulder, and Howey]{reniers2019review}
Jorn~M Reniers, Grietus Mulder, and David~A Howey.
\newblock Review and performance comparison of mechanical-chemical degradation models for lithium-ion batteries.
\newblock \emph{Journal of The Electrochemical Society}, 166\penalty0 (14):\penalty0 A3189--A3200, 2019.

\bibitem[Smith and Wang(2006)]{smith2006solid}
Kandler Smith and Chao-Yang Wang.
\newblock Solid-state diffusion limitations on pulse operation of a lithium ion cell for hybrid electric vehicles.
\newblock \emph{Journal of power sources}, 161\penalty0 (1):\penalty0 628--639, 2006.

\bibitem[Gasper et~al.(2022)Gasper, Collath, Hesse, Jossen, and Smith]{gasper2022machine}
Paul Gasper, Nils Collath, Holger~C Hesse, Andreas Jossen, and Kandler Smith.
\newblock Machine-learning assisted identification of accurate battery lifetime models with uncertainty.
\newblock \emph{Journal of The Electrochemical Society}, 169\penalty0 (8):\penalty0 080518, 2022.

\bibitem[Jiang et~al.(2021)Jiang, Gent, Mohr, Das, Berliner, Forsuelo, Zhao, Attia, Grover, Herring, et~al.]{jiang2021bayesian}
Benben Jiang, William~E Gent, Fabian Mohr, Supratim Das, Marc~D Berliner, Michael Forsuelo, Hongbo Zhao, Peter~M Attia, Aditya Grover, Patrick~K Herring, et~al.
\newblock Bayesian learning for rapid prediction of lithium-ion battery-cycling protocols.
\newblock \emph{Joule}, 5\penalty0 (12):\penalty0 3187--3203, 2021.

\bibitem[Dubarry and Anse{\'a}n(2022)]{dubarry2022best}
Matthieu Dubarry and David Anse{\'a}n.
\newblock Best practices for incremental capacity analysis.
\newblock \emph{Frontiers in Energy Research}, 10, 2022.

\bibitem[Dormann et~al.(2013)Dormann, Elith, Bacher, Buchmann, Carl, Carré, Marquéz, Gruber, Lafourcade, Leitão, Münkemüller, McClean, Osborne, Reineking, Schröder, Skidmore, Zurell, and Lautenbach]{dormann2013collinearity}
Carsten~F. Dormann, Jane Elith, Sven Bacher, Carsten Buchmann, Gudrun Carl, Gabriel Carré, Jaime R.~García Marquéz, Bernd Gruber, Bruno Lafourcade, Pedro~J. Leitão, Tamara Münkemüller, Colin McClean, Patrick~E. Osborne, Björn Reineking, Boris Schröder, Andrew~K. Skidmore, Damaris Zurell, and Sven Lautenbach.
\newblock Collinearity: a review of methods to deal with it and a simulation study evaluating their performance.
\newblock \emph{Ecography}, 36\penalty0 (1):\penalty0 27--46, 2013.

\bibitem[Cai et~al.(2018)Cai, Luo, Wang, and Yang]{cai2018fea_select}
Jie Cai, Jiawei Luo, Shulin Wang, and Sheng Yang.
\newblock Feature selection in machine learning: A new perspective.
\newblock \emph{Neurocomputing}, 300:\penalty0 70--79, 2018.
\newblock ISSN 0925-2312.

\bibitem[Hastie et~al.(2013)Hastie, Tibshirani, and Friedman]{element_stat_learn}
Trevor~J. Hastie, Robert~J. Tibshirani, and Jerome~H. Friedman.
\newblock \emph{The elements of statistical learning : data mining, inference, and prediction}.
\newblock Springer, New York, NY, 2nd edition, 2013.
\newblock ISBN 9780387848570.

\bibitem[Lake et~al.(2015)Lake, Salakhutdinov, and Tenenbaum]{lake2015human}
Brenden~M Lake, Ruslan Salakhutdinov, and Joshua~B Tenenbaum.
\newblock Human-level concept learning through probabilistic program induction.
\newblock \emph{Science}, 350\penalty0 (6266):\penalty0 1332--1338, 2015.

\bibitem[Pedersen et~al.(2019)Pedersen, Miller, Simpson, and Ross]{pedersen2019hierarchical}
Eric~J Pedersen, David~L Miller, Gavin~L Simpson, and Noam Ross.
\newblock Hierarchical generalized additive models in ecology: an introduction with mgcv.
\newblock \emph{PeerJ}, 7:\penalty0 e6876, 2019.

\bibitem[Bhattacharya et~al.(2018)Bhattacharya, Jaiswal, and Kumar]{bhattacharya2018faster}
Anup Bhattacharya, Ragesh Jaiswal, and Amit Kumar.
\newblock Faster algorithms for the constrained k-means problem.
\newblock \emph{Theory of computing systems}, 62\penalty0 (1):\penalty0 93--115, 2018.

\bibitem[Zhou and Howey(2022)]{zhou2022bayesian}
Zihao Zhou and David~A Howey.
\newblock Bayesian hierarchical modelling for battery lifetime early prediction.
\newblock \emph{arXiv preprint arXiv:2211.05697}, 2022.

\bibitem[Raj et~al.(2020)Raj, Wang, Monroe, and Howey]{raj2020investigation}
Trishna Raj, Andrew~A Wang, Charles~W Monroe, and David~A Howey.
\newblock Investigation of path-dependent degradation in lithium-ion batteries.
\newblock \emph{Batteries \& Supercaps}, 3\penalty0 (12):\penalty0 1377--1385, 2020.

\bibitem[Dahn et~al.(2012)Dahn, Smith, Burns, Stevens, and Dahn]{dahn2012user}
Hannah~M Dahn, AJ~Smith, JC~Burns, DA~Stevens, and JR~Dahn.
\newblock User-friendly differential voltage analysis freeware for the analysis of degradation mechanisms in li-ion batteries.
\newblock \emph{Journal of The Electrochemical Society}, 159\penalty0 (9):\penalty0 A1405, 2012.

\bibitem[Li et~al.(2019)Li, Colclasure, Finegan, Ren, Shi, Feng, Cao, Yang, and Smith]{li2019degradation}
Xuemin Li, Andrew~M Colclasure, Donal~P Finegan, Dongsheng Ren, Ying Shi, Xuning Feng, Lei Cao, Yuan Yang, and Kandler Smith.
\newblock Degradation mechanisms of high capacity 18650 cells containing si-graphite anode and nickel-rich nmc cathode.
\newblock \emph{Electrochimica Acta}, 297:\penalty0 1109--1120, 2019.

\bibitem[Keil and Jossen(2016)]{keil2016calendar}
Peter Keil and Andreas Jossen.
\newblock Calendar aging of nca lithium-ion batteries investigated by differential voltage analysis and coulomb tracking.
\newblock \emph{Journal of The Electrochemical Society}, 164\penalty0 (1):\penalty0 A6066, 2016.

\bibitem[Eddahech et~al.(2014)Eddahech, Briat, and Vinassa]{EDDAHECH2014determine_soh_cv}
Akram Eddahech, Olivier Briat, and Jean-Michel Vinassa.
\newblock Determination of lithium-ion battery state-of-health based on constant-voltage charge phase.
\newblock \emph{Journal of Power Sources}, 258:\penalty0 218--227, 2014.
\newblock ISSN 0378-7753.

\bibitem[Yang et~al.(2018)Yang, Xia, Huang, Fu, and Mi]{YANG2018online_soh_cv}
Jufeng Yang, Bing Xia, Wenxin Huang, Yuhong Fu, and Chris Mi.
\newblock Online state-of-health estimation for lithium-ion batteries using constant-voltage charging current analysis.
\newblock \emph{Applied Energy}, 212:\penalty0 1589--1600, 2018.
\newblock ISSN 0306-2619.

\bibitem[Wang et~al.(2019)Wang, Zeng, Guo, and Qin]{WANG2019soh_cv}
Zengkai Wang, Shengkui Zeng, Jianbin Guo, and Taichun Qin.
\newblock State of health estimation of lithium-ion batteries based on the constant voltage charging curve.
\newblock \emph{Energy}, 167:\penalty0 661--669, 2019.
\newblock ISSN 0360-5442.

\bibitem[Yuan and Yang(2019)]{yuan2019research}
Chunhui Yuan and Haitao Yang.
\newblock Research on k-value selection method of k-means clustering algorithm.
\newblock \emph{J}, 2\penalty0 (2):\penalty0 226--235, 2019.

\end{thebibliography}

\end{document}